\documentclass[10pt,twocolumn,letterpaper]{article}

\usepackage{iccv}              

\usepackage[accsupp]{axessibility} 
\usepackage{gensymb}
\usepackage{colortbl}
\usepackage[bottom]{footmisc}
\definecolor{iccvblue}{rgb}{0.21,0.49,0.74}
\usepackage[pagebackref,breaklinks,colorlinks,allcolors=iccvblue]{hyperref}

\def\paperID{2308} 
\def\confName{ICCV}
\def\confYear{2025}

\title{After the Party: Navigating the Mapping From Color to Ambient Lighting
}

\author{
Florin-Alexandru Vasluianu \quad 
Tim Seizinger  \quad
Zongwei Wu\thanks{Corresponding} \quad
Radu Timofte \\
Computer Vision Lab, CAIDAS \& IFI, University of W\"{u}rzburg \\
{\tt\small \{firstname.lastname\}@uni-wuerzburg.de}
}

\begin{document}
\maketitle
\begin{abstract}
Illumination in practical scenarios is inherently complex, involving colored light sources, occlusions, and diverse material interactions that produce intricate reflectance and shading effects. However, existing methods often oversimplify this challenge by assuming a single light source or uniform, white-balanced lighting, leaving many of these complexities unaddressed. In this paper, we introduce CL3AN, the first large-scale, high-resolution dataset of its kind designed to facilitate the restoration of images captured under multiple Colored Light sources to their Ambient-Normalized counterparts. Through benchmarking, we find that leading approaches often produce artifacts—such as illumination inconsistencies, texture leakage, and color distortion—primarily due to their limited ability to precisely disentangle illumination from reflectance. Motivated by this insight, we achieve such a desired decomposition through a novel learning framework that leverages explicit chromaticity-luminance components guidance, drawing inspiration from the principles of the Retinex model. Extensive evaluations on existing benchmarks and our dataset demonstrate the effectiveness of our approach, showcasing enhanced robustness under non-homogeneous color lighting and material-specific reflectance variations, all while maintaining a highly competitive computational cost. The benchmark, codes, and models are available at www.github.com/fvasluianu97/RLN2.
\end{abstract}

\begin{figure}
    \centering
    \includegraphics[width=\linewidth]{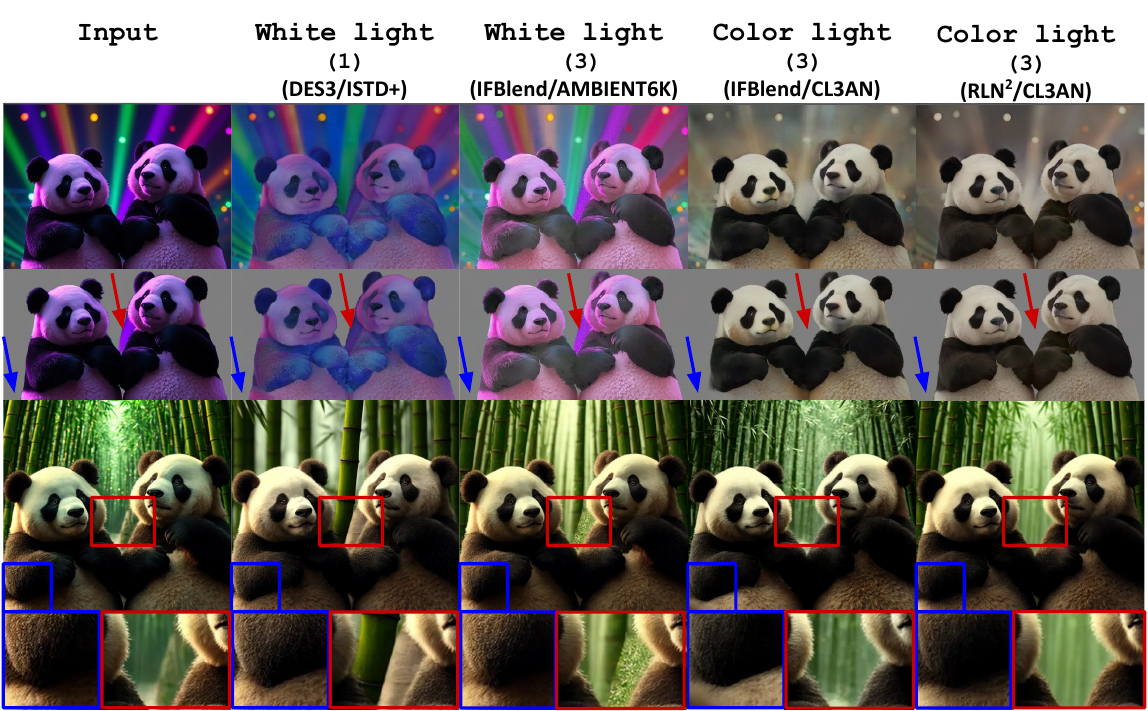}
    \caption{Effective deployment of Ambient Lighting Normalization (ALN) as a pre-processing step in Neural Image Editing \cite{fu2023guiding}. We compare the effectiveness of our proposed RLN$^2$ against state-of-the-art models in fields like Shadow Removal (DES3 \cite{jin2022des3}), or ALN (IFBlend \cite{vasluianu2024towards}), trained on settings including single (1) \cite{wang2018stacked} or multiple (3)  white-domain-tuned \cite{vasluianu2024towards} or color light sources. }
    \label{fig:teaser_dia}
    \vspace{-5mm}
\end{figure}

\section{Introduction}
\label{sec: intro}

The goal of Ambient Lighting Normalization is recovering an ambient-lit image, under uniform lighting distribution, regardless of the lighting scenario characteristic to the depicted contents \cite{vasluianu2024towards}. 
The ability to recover ambient-normalized images—free from spatially varying illumination artifacts—is critical for robust visual understanding \cite{minaee2021image, ma2024u,jain2023oneformer}. In both practical imaging scenarios \cite{ignatov2020replacing, huang2018range} and the rapidly evolving domain of image generation and synthesis \cite{rao2024lite2relight, yoon2024generative}, illumination always involves complex interactions of multiple colored light sources, occlusions, and material properties, resulting in intricate shadows, glare, color shifts, and texture distortions. 
Lighting and material characteristics are the driving factors for explaining the manifestation of such artifacts, thus explaining the need for data-level representativity prior to proposing image restoration solutions aiming for improved degradation robustness or restoration performance. 

While a wide range of applications require illumination-invariant representations \cite{ humeau2019texture, barron2014shape, wan2024interpretable, sarlin2020superglue}, existing normalization methods struggle to disentangle these intertwined factors \cite{toschi2023relight, jin2025neural, fu2023guiding, yoon2024generative}, as illustrated in \cref{fig:teaser_dia}. Even if data contributions further away from the natural light in ISTD+ \cite{wang2018stacked, le2021physics}, or multiple white-light-tuned light sources from AMBIENT6K \cite{vasluianu2024towards} can improve general effectiveness, model design can further drive robustness. 

This limitation stems primarily from oversimplified assumptions, such as single white light sources \cite{wang2018stacked, vasluianu2023wsrd} or homogeneous color fields \cite{wang2018stacked, qu2017deshadownet, zhang2020portrait}, which fail to capture the richness of complex illumination. A key technical challenge lies in the lack of comprehensive datasets that adequately represent the complexity of real-world lighting, especially outside the realm of natural white-color-tuned lighting. Prior efforts in shadow removal \cite{wang2018stacked, vasluianu2023wsrd}, color constancy \cite{foster2011color, finlayson2001color, barron2015convolutional}, or color correction \cite{mantiuk2009color, guerreiro2023pct, jeon2024low, afifi2020deep, weng2005novel} address isolated aspects of illumination but fall short of modeling multi-light interactions. Recent attempts \cite{kim2021large, vasluianu2024towards} introduce variable-intensity lighting datasets but lack chromatic diversity and material-specific reflectance variations. Furthermore, wide access to ambient-lit images outside the scope of scene-rendering software remains a challenge.  As a result, learning-based methods often produce artifacts.

In this paper, we address these limitations through two core contributions. First, we introduce CL3AN, the first large-scale dataset capturing high-resolution scenes under multiple colored directional lights paired with ambient-lit references. Each scene triplet includes variations beyond white-balanced lighting, colored lighting, and a diffuse-lit ground truth, enabling systematic study of illumination effects like self-shadows, highlights, glare areas, color spill or saturation, and other material-dependent artifacts. By benchmarking leading image restoration methods \cite{guo2025mambair, cai2023retinexformer, li2023grl, vasluianu2024towards}, we observe that these models frequently suffer from critical shortcomings, such as the conflation of chromatic variations caused by illumination, texture leakage, inconsistent partial normalization, etc. 

We identify the fundamental limitation of existing methods as their inability to systematically disentangle reflectance and illumination—a critical flaw exacerbated by the lack of explicit guidance for separating chromaticity and luminance under multi-colored lighting. Most current approaches \cite{zamir2022restormer, li2023grl, guo2025mambair} rely on single-stream, end-to-end feature modeling, which implicitly maps input images to ambient-normalized counterparts without incorporating physical awareness of light-material interactions. While some recent works attempt to address this by introducing pixel- and frequency-domain supervision \cite{cui2023selective, vasluianu2024towards}, more flexible towards reflectance- and chromaticity-information extraction, they remain limited in learning appropriate frequency filters — a challenging task without explicit guidance — resulting in suboptimal performance under the challenging multi-colored lighting conditions.

In this paper, from a novel perspective, we propose \textbf{RLN$^2$}, a Retinex-inspired framework that introduces parallel, physics-grounded streams to independently model reflectance (material properties) and illumination (lighting geometry/color). We observe that these intrinsic specificities are naturally encoded in the cylindrical representation of the input image, where the Hue (H) and Saturation (S) maps align with the observed chromaticity under the effect of the incoming light, while the Value (V) map provides strong priors for materials-specific reflectance. By embedding these domain-specific clues — such as reflectance information entropy (V), and smoothness constraints (H, S) — into feature refinement, our framework achieves precise decomposition, effectively eliminating undesirable artifacts while preserving fine-grained details.

\begin{figure*}
    \centering
    \includegraphics[width=\linewidth]{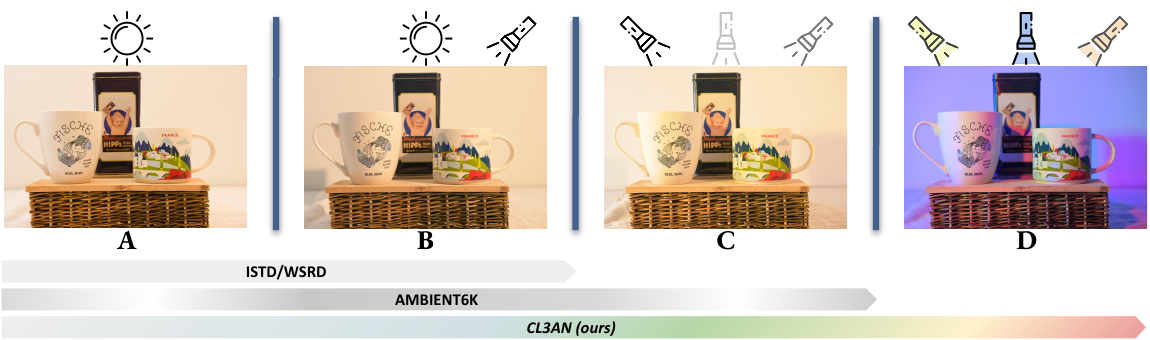}
    \caption{A comparison between the lighting scenarios represented in datasets targeting different applications of Ambient Lighting Normalization (ISTD \cite{wang2018stacked}, WSRD\cite{vasluianu2023wsrd}, or AMBIENT6K \cite{vasluianu2024towards}). In \textbf{(A)}, ambient lighting is used to acquire a uniformly-lit reference image. In \textbf{(B)}, direct lighting is used for shadow casting, while constraining consistency to the reference image in shadow-free segments. In \textbf{(C)}, more directional lights are added to the direct lighting system. In our proposed setup \textbf{(D)}, the direct lighting setup is based on RGB lights, dropping the color consistency constraint, and showcasing complex material-light interactions under various conditions. }
    \label{fig:samples-light}
    \vspace{-3mm}
\end{figure*}

\noindent Our contributions can be summarized as follows:
\begin{itemize}
    \item The introduction of the \textbf{CL3AN}, a first-of-its-kind dataset enabling the study of ambient lighting normalization under multiple RGB lights;
    \item The extensive study of a wide range of models, seminal works in various image restoration tasks, when trained and tested on the \textbf{CL3AN} dataset;
    \item We propose \textbf{RLN$^\text{2}$}, a robust algorithm built on the principle of simultaneous image reflection and lightness components compensation to recover an ambient lit image, characterized by uniform lighting distribution.  \textbf{RLN$^\text{2}$} is proposed as a strong baseline for the ambient lighting normalization task, achieving state-of-the-art results on both the public AMBIENT6K dataset and the proposed \textbf{CL3AN} benchmark. 
\end{itemize}

\section{Related Work}
\label{sec: related}
This work expands Ambient Lighting Normalization to multicolor direct lighting, addressing previously unexplored effects appearing in cluttered scenes, and analyzing shadow interactions and highlights appearing under multicolor direct lighting.

\noindent \textbf{Lighting and Color:} 
Color analysis is central to image processing and understanding, with various downstream tasks such as facial recognition \cite{xie2006efficient, 5658185,takano2007rotation}, or tracking and matching applications \cite{lahiri2021lipsync3d, fang2022alphapose, sarlin2020superglue, bansal20212d, 10054148} relying on the color similarity principle, thus benefiting from increased appearance robustness, especially under challenging lighting. 

Image Quality Assessment (IQA) methods \cite{agnolucci2024arniqa, yang2022maniqa} show a strong correlation between details-abundance and image quality, motivating the focus on professional scene lighting rigs in all expert image photography applications. 
Given the scarcity of such equipment, alternative strategies \cite{barron2015convolutional, afifi2020deep, afifi2021cross, li2022mimt} are developed around image sensor sensor properties and cross-device constancy \cite{gijsenij2011color, boyadzhiev2012user}.  

Naturally, registered image stacks either for tasks such as multiple view geometry estimation \cite{Zhu_2022_CVPR, 10.1007/978-3-642-15561-1_27} or different image stitching applications require consistent color and content appearance \cite{szeliski2022image, lai2019video, cheng2022inout, araujo2023towards}.
In such applications, the effort is focused on consistent lighting in the image acquisition, rather than post-processing correction. In-the-wild applications focusing on widely available amateur images \cite{szeliski2022image, gupta2023asic, 10.1007/978-3-642-15561-1_27} document extensive data curation strategies prior to model estimations.

Later applications, given the newly-found popularity of neural 3D modeling, approach scene relighting as part of their rendering pipeline \cite{rudnev2022nerf, toschi2023relight}. 
Alternatively, approaches based on complex generative image models \cite{ponglertnapakorn2023difareli, jin2025neural} successfully tackle lighting manipulation, but at a high computational cost.
Unfortunately, the limited availability of real-world ambient lit images, primarily due to the complexity of an ambient lighting setup, explains the difficulty in addressing real-world lighting normalization, even with the obvious availability of large-complexity models.   

Reliance on natural light for ambient-lit images proved a viable option, but hindered the variety of possible observed degradations, especially in terms of chromaticity shifts. 
Unfortunately, the large number of factors involved in light occlusion interaction greatly constrains a representative simulation in general settings.
While ISTD/ISTD+ \cite{wang2018stacked, le2021physics}, SRD \cite{qu2017deshadownet} or USR \cite{hu2019mask} show exclusively flat surfaces under the influence of natural light, the influence of object self-shadows was just recently represented in WSRD \cite{vasluianu2023wsrd}, only possible under a laboratory setup. 

Recently, multiple source lighting gained in data-level representation.
For example, in LSMI \cite{kim2021large}, the authors target White-Balance correction robustness under natural and artificial lighting. 
Alternatively, effects characteristic to multiple illuminants were captured in works such as \cite{hui2019learning, murmann2019dataset}, or in the context of flash photography \cite{hui2018illuminant}. 
However, a uniformly-lit ambient normalized image can not be acquired in these settings.

Alternatively, the introduction of AMBIENT6K \cite{vasluianu2024towards} coupled multiple light sources of varying intensity and orientation for complex interactions, while also providing an ambient-lit image based on an alternative lighting setup. 
Unfortunately, the strong connection between the operating parameters set for the direct and ambient lighting systems limits the study to exposure correction for white-aligned scene lighting (see \cref{fig:samples-light} \textbf{(A)}, \textbf{(B)}, and \textbf{(C)}). 

\noindent \textbf{Image Restoration (IR):} 
Joint color and exposure correction, while restoring details lost to occlusion or saturation, defines Ambient Lighting Normalization as an Image Restoration task.

Various Convolution Neural Networks (CNNs) \cite{7839189, jing2021hinet, chen2022simple, zamir2021multi} became strong baselines for the family of IR tasks. 
The later introduction of the attention mechanism \cite{vaswani2017attention} in image-based applications \cite{dosovitskiy2020image} enabled consistent performance advantages in the wide IR field. 
Focusing on different feature fusion techniques through specifically tailored attention mechanisms, based on the IR task particularities, defines the success of multiple models \cite{liang2021swinir, wang2022uformer, chen2022simple, zamir2022restormer, 10462902, li2023grl} easily becoming benchmark solutions for various IR problems. 
Separate refinement for image reflectance and lightness under the Retinex model \cite{6872f11f-4c61-3054-ba57-067328d65b02} proved particularly effective in low-light image enhancement (LLIE) \cite{wei2018deep, zhu2020zero, cai2023retinexformer, xu2024cretinex}, or image shadow removal \cite{huang2024image}.  

Global range dependencies benefits drive the development of State Space Models (SSM) \cite{gu2023mamba}, with promising results in general IR \cite{guo2025mambair}. 
Alternatively, global knowledge can be leveraged through the integration of alternative frequency image representations in IR solutions.
Solutions such as FFTFormer \cite{kong2023efficient}, DCFormer \cite{li2023discrete},  SFNet \cite{cui2023selective}, DW-GAN \cite{Fu_2021_CVPR}, and IFBlend \cite{vasluianu2024towards} successfully extract priors based on Fourier Transform, Discrete Cosine Transform, or other  Wavelet Transform variants. 

Finally, diffusion-based iterative recurrent solutions  \cite{ho2020denoising} started to mature as benchmark solutions in Image Restoration  \cite{xia2023diffir, zhu2023denoising, luo2023refusion, wang2024exploiting}, especially in terms of perceived image quality. 
However, their characteristic increased computational effort hinders their deployment.   

\begin{figure}[t!]
    \centering
    \includegraphics[width=\linewidth]{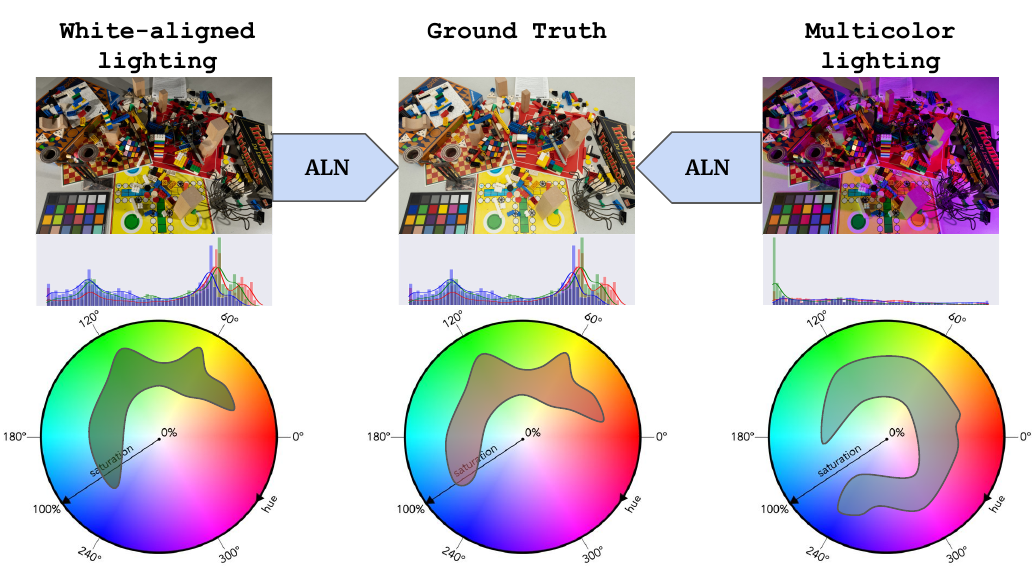}
    \caption{A graphical representation of the Ambient Lighting Normalization transformation in the context of the white-aligned lighting, compared to the challenges posed by the introduction of a multicolor direct lighting system. The RGB image histogram, coupled with the Hue-Saturation rough projection is provided as a reference. Note the complexity of the change in image statistics.}
    \label{fig:dataset_fig}
    \vspace{-4mm}
\end{figure}

\section{Method}
\label{sec: method}
The core of our work is extending the study of Ambient Lighting Normalization to direct color lighting. 

\cref{fig:dataset_fig} compares the properties of a scene depicted under multi-source color direct lighting, and an equivalent image under white-aligned direct lighting. The goal to recover the ambient-lit reference image. 
Provided statistics, such as the RGB image histogram, and the Hue/Saturation footprint showcase the complexity of the ALN transformation, especially increasing with the introduction of variable intensity multicolor direct light sources. 
Naturally, the most notable difference comes from the color hue deviation. 
Materials adopt the color of incoming light, with the reflected light wavelength depending on the reflectance defining the represented materials, and the incoming light properties. 

Furthermore, the complex geometry of the scene drives light occlusions, with numerous self-shadows at the image level. 
In this case, given the non-uniform color distribution, the color of the observed shadows depends on the local lighting availability and their light source-defining properties. 
A successful restoration of the ambient-lit image can not be performed without a full understanding of both scene geometry (for local exposure compensation) and lighting system configuration (for chromaticity uniformization, given the arithmetic of the locally available light radiation given the material properties).

\begin{table*}
\centering
\setlength{\tabcolsep}{3.5pt}
\vspace{-3mm}
\resizebox{\linewidth}{!}
{
\begin{tabular}{ccccc|cccc|cccc}
\toprule
\rowcolor{gray!20}
\multicolumn{5}{c|}{\textbf{Camera parameters}} & \multicolumn{4}{c|}{\textbf{Diffuse Lighting}} & \multicolumn{4}{c}{\textbf{Direct Lighting}} \\
\midrule
Exposure & Aperture & ISO & White & Focal & \# Active & White light & White light & Intensity & \# Active & White light & Intensity & Color Hue \\
time (s) & & & balance & dist. (mm) & lights & temp. (K) & temp. range & & lights& temp. (K) & range & range \\
\midrule
1/60 & f11 & 100 & 6400K & 50 & 5 & 6400 & $\pm$ 5\% & 100\% & $\ge 1$ & 6400 & 30-100\% & 0$\degree$-360$\degree$ \\
\bottomrule
\end{tabular}
}
\vspace{-2mm}
\caption{\textbf{Dataset capturing parameters} regarding the setup of the camera, the diffuse lighting system, and the direct lighting hardware.}
\label{tab:dataset-comparison}
\vspace{-4mm}
\end{table*}

\subsection{CL3AN}
\label{subsec: dataset}
\textbf{CL3AN} represents an image collection representing \textbf{Tri}plets consisting of images acquired under both \textbf{C}olor and White \textbf{L}ighting, and \textbf{A}mbient \textbf{N}ormalized images, under uniformly distributed diffuse light.

As introduced in \cref{fig:samples-light}, \textbf{CL3AN} is the only dataset providing equivalent representations of the same scene under white-aligned direct lighting (setups \textbf{(B)}, \textbf{(C)}), varying intensity multicolor lighting (setup \textbf{(D)}), while also providing an ambient-lit reference image (setup \textbf{(A)}). 
This makes \textbf{CL3AN} a \emph{first-of-its-kind} dataset. 

The capturing system relies on a professional photography studio setup, in which a dark room is used to ensure that all image-capturing parameters become trackable and controllable. 
Two separate scene lighting systems are concurrently developed, optimizing for uniform diffuse light distribution, respectively direct lighting, being triggered only for the acquisition of the corresponding samples. 
Note that directional lights are fully controllable, so they can be tuned to the white natural light domain, producing images similar to those available in datasets such as AMBIENT6K \cite{vasluianu2024towards}.   

The images are captured with a tripod-mounted and remotely operated Canon R6 MKII camera, at constant focal range. 
This leads to a perfect pixel alignment between the images characteristic to each triplet. 
For consistency and traceability, the camera-related parameters are kept fixed during the image-capturing process.
All hardware parameters relevant to data acquisition are documented in \cref{tab:dataset-comparison}.

For the direct lighting system, three programmable mobile RGB lights are used to capture scene variants under variable light color, intensity, position, and orientation. 
The direct consequence of this set of parameters is a non-homogeneous distribution of the incoming light color across the scene, implying that global color correction can not be used for restoration in this scenario (see \cref{fig:samples-light} and \cref{fig:dataset_fig}). 

The diffuse ambient lighting system is composed of five white softbox lights, covering all sides of the cubic acquisition environment, except the base, which is the surface supporting the captured contents.  
Thus, the possibility for self-shadows in the ambient-lit images is almost fully mitigated, being furthermore enforced through geometry-based object filtering.  Calibration over a 24-sample color checker under full white light intensity enables the alignment of both lighting setups, by determining the set of operating parameters producing a similar colors under optimal exposure.  

\textbf{CL3AN} is a scene-based dataset, covering 105 cluttered scenes representing an extensive set of materials, both conductors and dielectrics, with varying transparency, color, and surface roughness.
The data is available as both RAW and RGB images, at 24 MP, the camera sensor resolution. 
Data splitting is performed based on the represented contents, with 10 scenes used exclusively for testing, and 10 scenes for validation.
Both testing and validation splits represent objects unseen during training, but under lighting conditions consistent with the training split. 
Additional metadata listing the objects within a scene is provided.
The training split consists of a number of 3667 samples, with 437 images used for validation and 431 for testing.

\subsection{RLN$^2$}
\label{subsec: model}

\cref{fig:samples-light} and \cref{fig:dataset_fig} show the challenging environment posed by the introduction of multicolor direct lighting in the ALN field.
Dramatic changes can be observed when mapping from the input image to the ambient-lit reference, especially in terms of observed chromaticity. 
Moreover, light occlusion by the many rough surfaces corresponding to the represented objects drives a wide variety of intricate effects, with a strong information loss. 
Learning a robust ALN mapping requires a robust design, able to disentangle significant information compensation in confined image segments from the large magnitude chromaticity shifts. 

\begin{figure}[t!]
    \centering
    \includegraphics[width=0.9\linewidth]{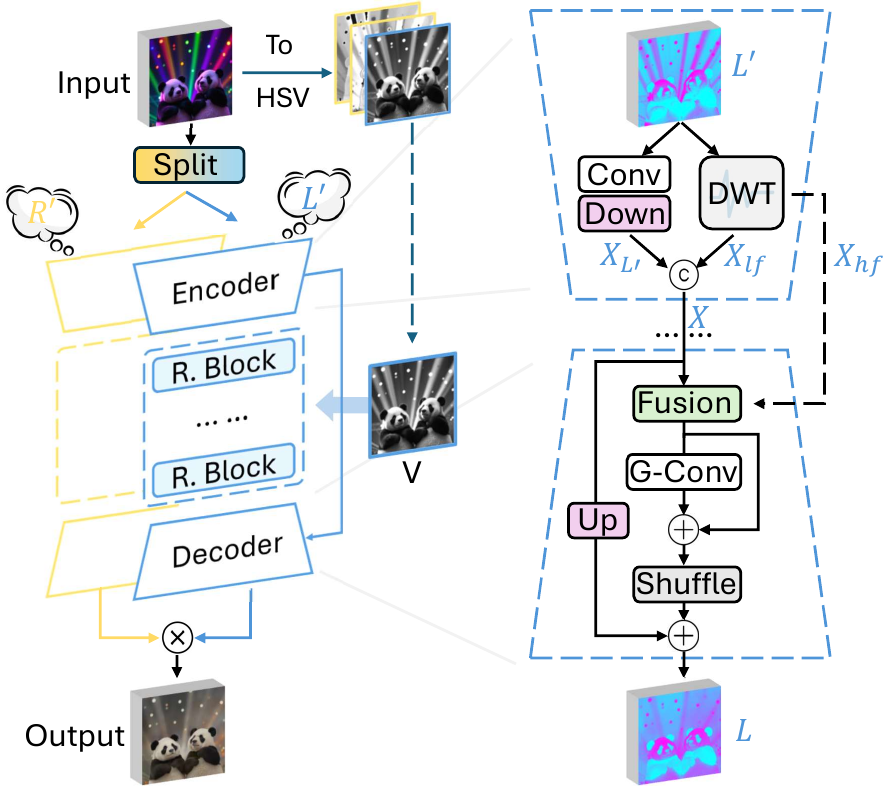}
    \vspace{-4mm}
    \caption{A graphical representation of the RLN$^2$ architecture, including detailed representations for an Encoder (E) - Decoder (D) blocks pair on the Luminance L-branch. The Reflectance R-branch design is similar, being omitted for clarity.}
    \label{fig:model_dia}
    \vspace{-4mm}
\end{figure}

Given these observations, problem sub-division is the core of our proposed design RLN$^2$, aiming for \underline{joint} independent compensation in both Image \underline{R}eflectance and \underline{L}uminance components.   
Under the Retinex theory \cite{6872f11f-4c61-3054-ba57-067328d65b02}, a given image $I$ can be decomposed as a product of two multiplicative terms, representing image illumination or lightness $L$, and the material specific image reflectance $R$.

Thus, an image pair $(I, I')$ in which $I'$ is the degraded version of the reference image $I$, can be represented by the component set $\{(L, R), (L', R')\}$, in line with \cref{eqn:retinex}. 
\begin{equation}
    \begin{split}
        I(x, y) &= L(x, y) \cdot R(x, y) \\
        I'(x, y) &= L'(x, y) \cdot R'(x, y)
    \end{split}
    \label{eqn:retinex}
\end{equation}

Therefore, our goal is to determine the pair $(\overline{L}, \overline{R})$ in which $\overline{L}$ restores the details lost due to inadequate lighting, while $\overline{R}$ aligns the input chromaticity with the white-aligned ambient lit setup.
\begin{equation}
    \begin{split}
        L(x, y) &= L'(x, y) +\overline{L}(x, y) \\
        R(x, y) &= R'(x, y) +\overline{R}(x, y)
    \end{split}
    \label{eqn:residuals}
\end{equation}

 \cref{fig:model_dia} provides a graphical representation of the proposed RLN$^2$, in which the $(R', L')$ components of the input $I'$ are subjected to joint independent compensation, while the alternative HSV image representation information is used for feature refinement.  
 Aligned with the properties of the HSV image representation, which codes color in a cylindrical coordinate system, in which the Value (V) channel is similar to explicit illumination intensity, and the Hue-Saturation (HS) represents color tones, the V channel is used as guidance on the L-branch of our model, while the HS pair drives feature refinement on the R-branch of RLN$^2$. 

At the encoder level, the information goes through successive decomposition, in which Haar-DWT is used for spectral dissemination, alongside RGB feature downsampling.
The dual-domain information will be altered in the feature refinement stage, at the \emph{R.block} level, through the Cross-Domain Feature Fusion Attention module (CDFFA). 

\begin{figure}
    \centering
    \includegraphics[width=0.4\linewidth]{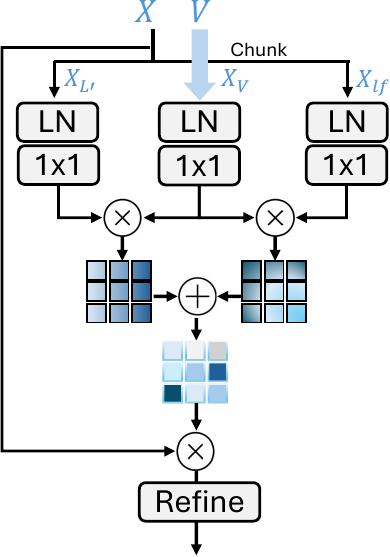}
    \vspace{-4mm}
    \caption{A detailed  Reflectance/Luminance Refinement Block (R.Block), including the CDFFA attention module.}
    \label{fig:attn_dia}
\end{figure}

Here, the low-frequency information $X_{lf}$, and the RGB extracted features $X_{L'}$, projected to an n-dimensional space, are then subjected to relevance enhancement, through a set of parameters based on dot product vector similarity (see \cref{fig:attn_dia}) to features extracted from the guidance component. 
The V channel controls the estimation of the $\overline{L}$ residual, while H and S drive the reflectance estimation $\overline{R}$. 

\begin{equation}
    \begin{split}
        X_{L'} &= proj_n(X_{L'})\\
        X_{lf} &= proj_n(X_{lf})\\
        \alpha_{i} &= sim(X_{L', i}, X_{V, i}) \\
        \beta_{i} &= sim(X_{lf, i}, X_{V, i})
    \end{split}
\end{equation}
The set of similarity coefficients is used to define a relevance probability distribution, then used to enhance the relevant information by filtering outlier features. 
The enhanced information is further refined before output. 
\begin{equation}
    \begin{split}
        \alpha &= softmax(\alpha) \\
        \beta &= softmax(\beta) \\
        X_{out} &=  \{ \alpha_i \cdot X_{L', i}\}_{i=0}^n + \{\beta_i \cdot X_{lf, i}\}_{i=0}^n
    \end{split}
\end{equation}

 At the decoder stage, the refined information is concatenated to the high-frequency spectral information from the corresponding encoder block.  
Moreover, an ImageNet \cite{10.5555/2999134.2999257} pretrained ConvNeXt-based feature extractor module \cite{liu2022convnet}, provides wide contextual information in feature fusion. 
A cross-attention module handles information exchange between the wide context information and the dual-domain features refined in the previous stage. 

 For a fair comparison, at this level, RLN$^2$ benefits from priors identical to \cite{vasluianu2024towards}, even if the implemented strategies are different.
 After linear projections, irrelevant features are suppressed through Channel Attention \cite{woo2018cbam}, being further refined for Reflectance/Luminance compensation \cite{yu2024mambaout}.

\noindent \textbf{Ablative study}:  
\cref{tab:ablation-alnd-test} compares the standard RLN$^2$ configuration, as represented in \cref{fig:model_dia}, to three similar variants in which the control mechanism implemented in the middle stage is modified. 
Reducing the \emph{R.block} to simple feature refinement by removing any control mechanism shows an important performance deficit, consistent in image restoration fidelity and perceptual quality. 
Alternatively, using the RGB domain images, passed through a dynamical filter and then split based on the filtered output and its alternative residual shows a slight improvement, at a relatively low induced computational cost. 

Task-aligned image representations, with explicit coding of image luminance and chromaticity, prove rich in relevant information, adding only the image conversion cost to the general computation. 
Focusing on relevant feature enhancement through the CDFFA introduction further increases model robustness and restoration performance. 

\noindent\textbf{Implementation:} RLN$^2$ was developed in PyTorch, and the documented experiments were performed on up to three NVIDIA L40s 48 GB VRAM GPUs running a CUDA 12.6 environment. 
The optimization is split into multiple stages, with the resolution of the training patches gradually increasing as the training procedure progresses.  
The optimizer used was Adam, with the learning rate $lr=0.0002$, and the loss optimized was the L1 distance. A cosine learning rate scheduler was used, with two oscillation periods. 
Optimization deploys gradient clipping, with a standard value of 0.01 under the L1 norm. 
Further details regarding the training procedure description can be found in the attached supplementary material.   
\begin{table}[t]
\centering

\resizebox{\linewidth}{!}
{%
\begin{tabular}{lcccccc}
\toprule
Model & Fusion & Guidance & MACs (G.) & \text{PSNR}$\uparrow$ & \text{SSIM}$\uparrow$ & \text{LPIPS}$\downarrow$ \\ \midrule
RLN$^2$ & \emph{None} & \emph{None} & 20.95     & 19.892                & 0.689                & 0.271                         \\
RLN$^2$  & \emph{concat.} & \emph{RGB}   & 23.67    &  20.018               &  0.703                & 0.233                         \\ 
RLN$^2$  & \emph{concat.} & \emph{L*a*b*} & 23.68     & 20.149        & 0.724                 & 0.226                \\ 
RLN$^2$  & \emph{concat.}  & \emph{HSV} & 23.68     & 20.216 & 0.731 & 0.220       \\ 
\midrule
RLN$^2$  & \emph{(CDFFA)}  & \emph{HSV}  & 22.72     & \textbf{20.523} & \textbf{0.746} & \textbf{0.208}   \\
\bottomrule
\end{tabular}%
}
\vspace{-2mm}
\caption{Inner-stage guidance feature fusion ablation study conducted on the testing split of the CL3AN dataset. }
\label{tab:ablation-alnd-test}
\vspace{-6mm}
\end{table}

\section{Results}
\label{sec: results}

\begin{figure*}
    \centering
    \setlength{\tabcolsep}{1pt}
    \renewcommand{\arraystretch}{0.7}
    \def\widthcomp{0.1639}
    \begin{tabular}{cccccc}
         Input    & HINet \cite{jing2021hinet}        &   NAFNet \cite{chen2022simple}    & IFBlend \cite{vasluianu2024towards}   & RLN$^2$-Lf \emph{(ours)}  & Ground Truth  \\
\includegraphics[width=\widthcomp\linewidth]{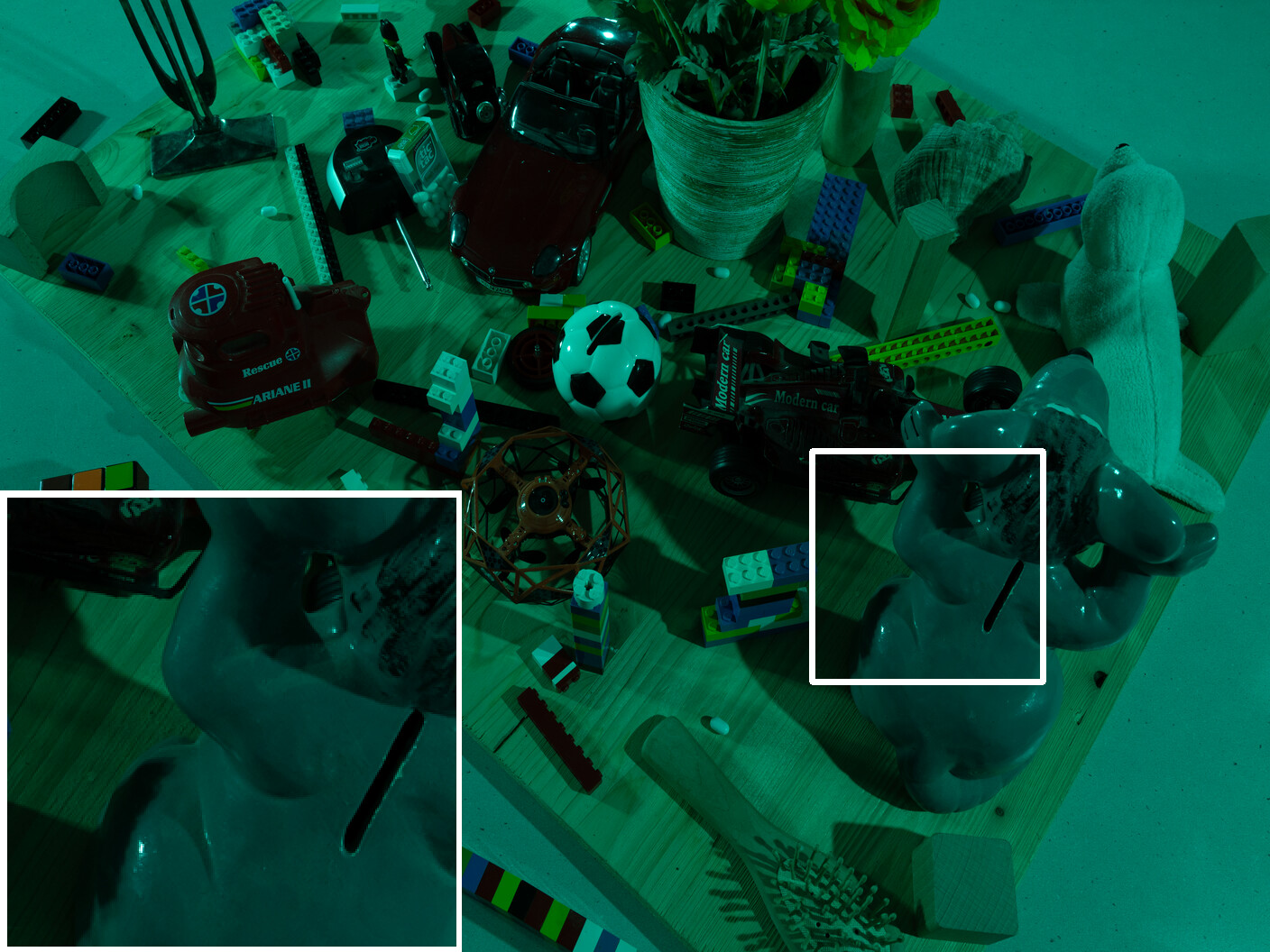} &
\includegraphics[width=\widthcomp\linewidth]{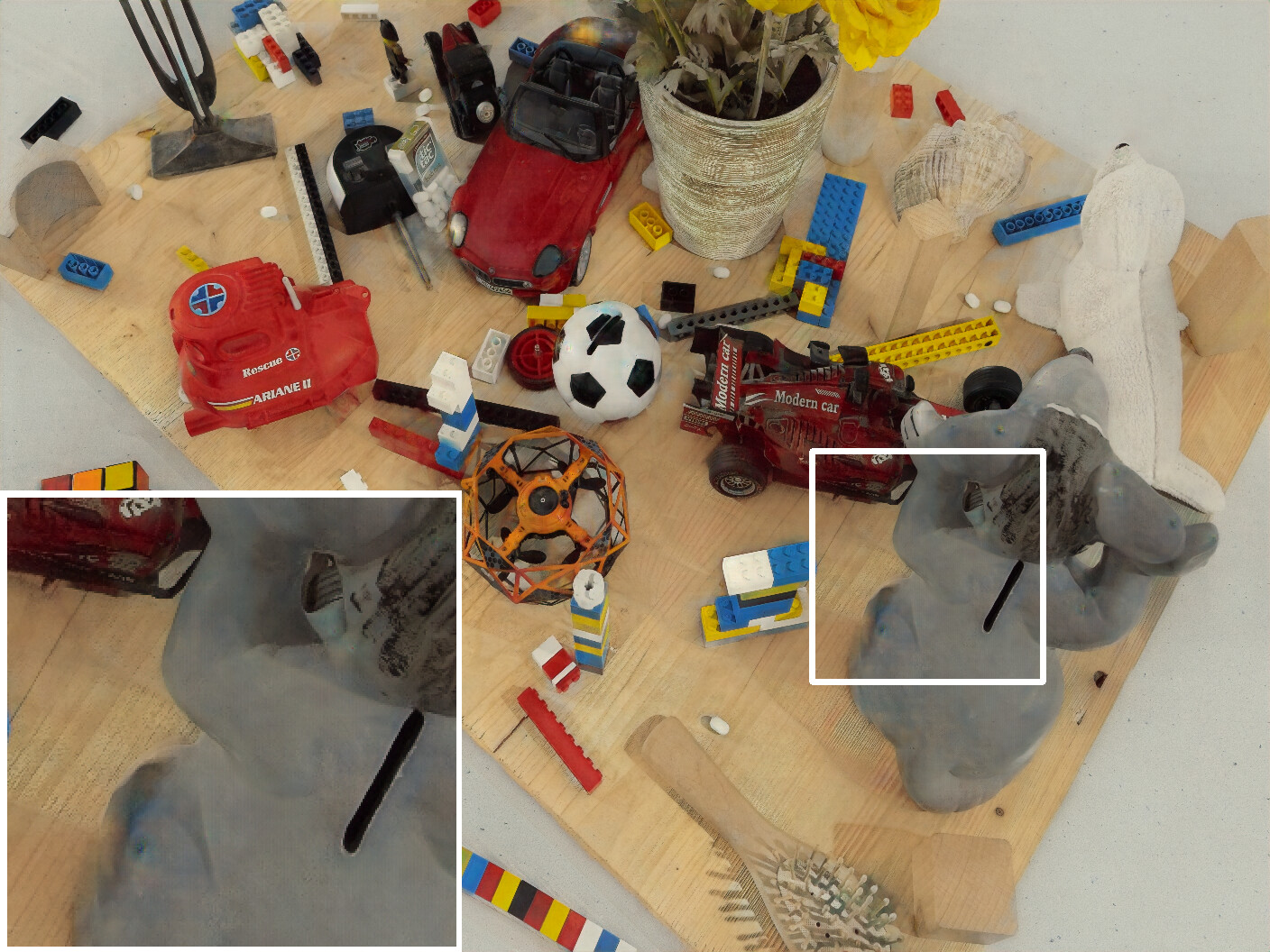} &
\includegraphics[width=\widthcomp\linewidth]{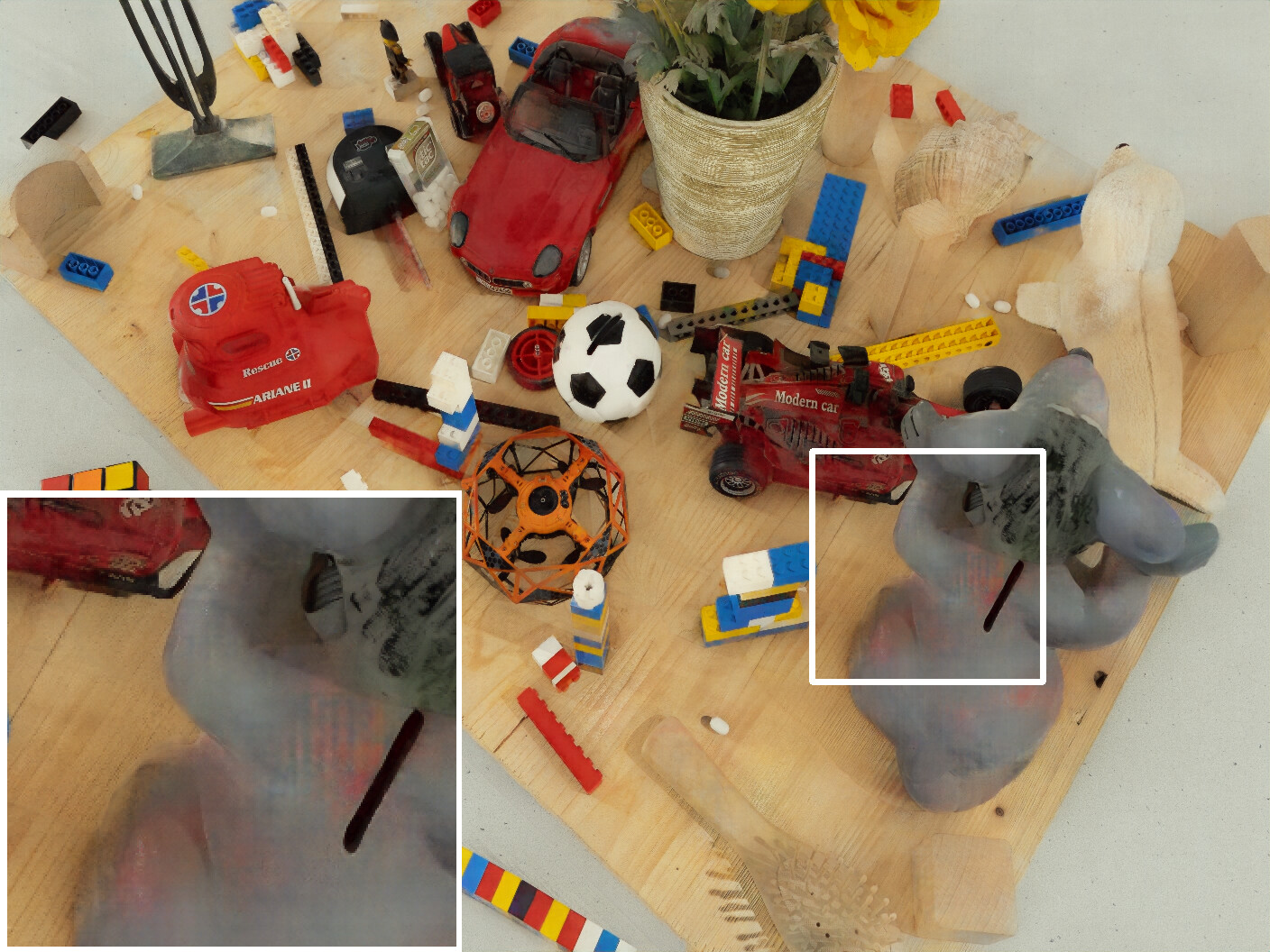} &
\includegraphics[width=\widthcomp\linewidth]{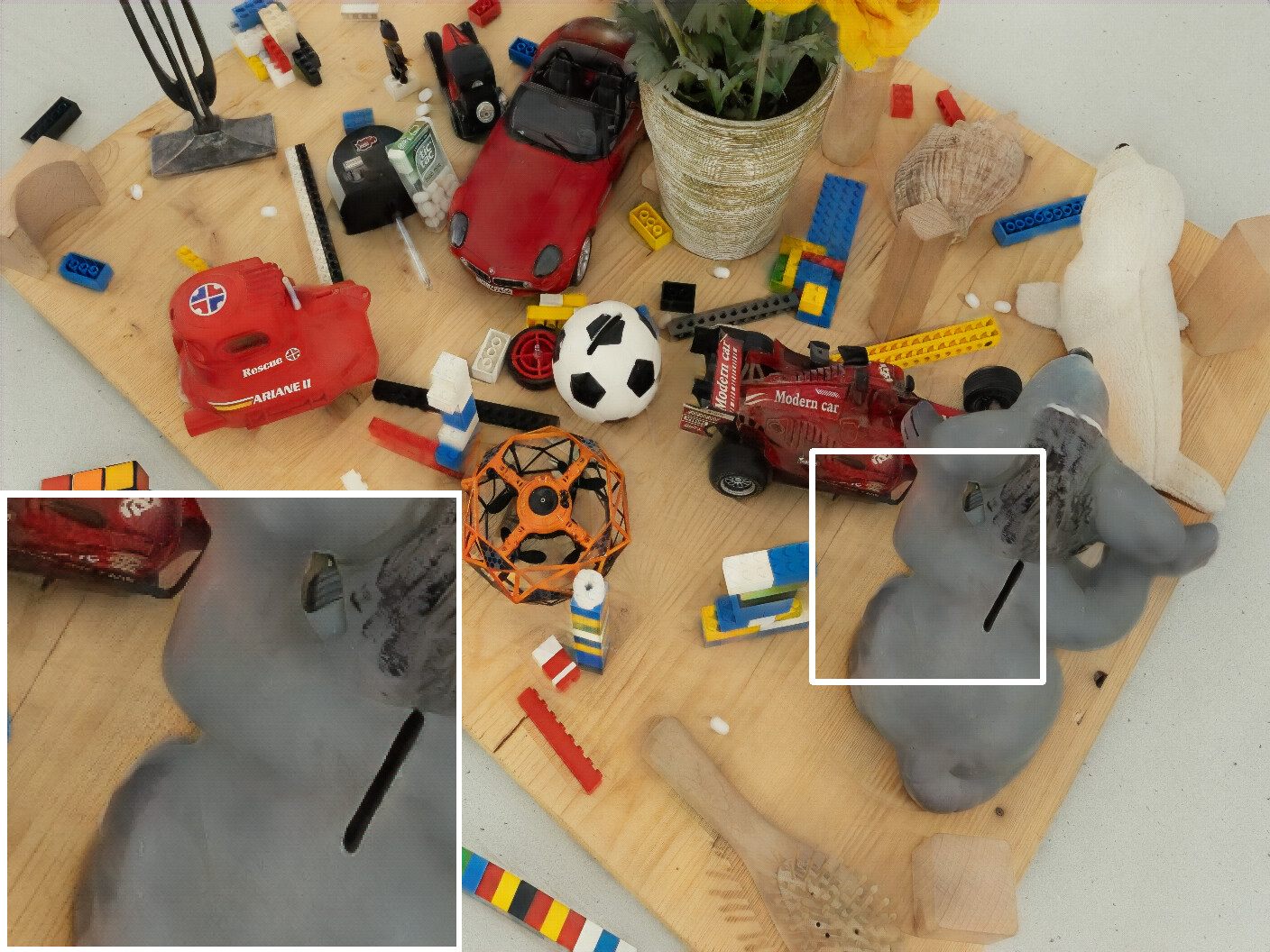} &
\includegraphics[width=\widthcomp\linewidth]{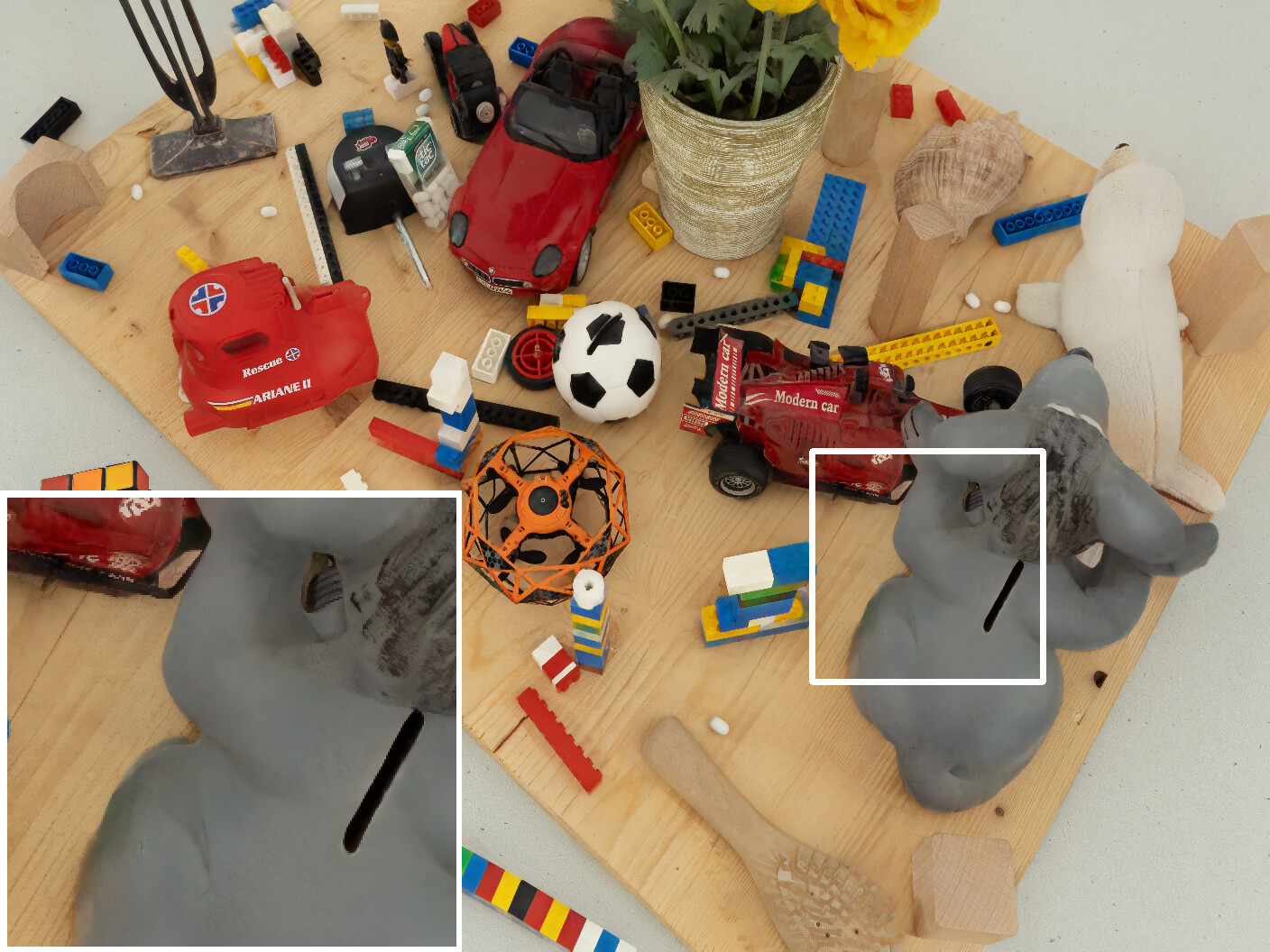} & 
\includegraphics[width=\widthcomp\linewidth]{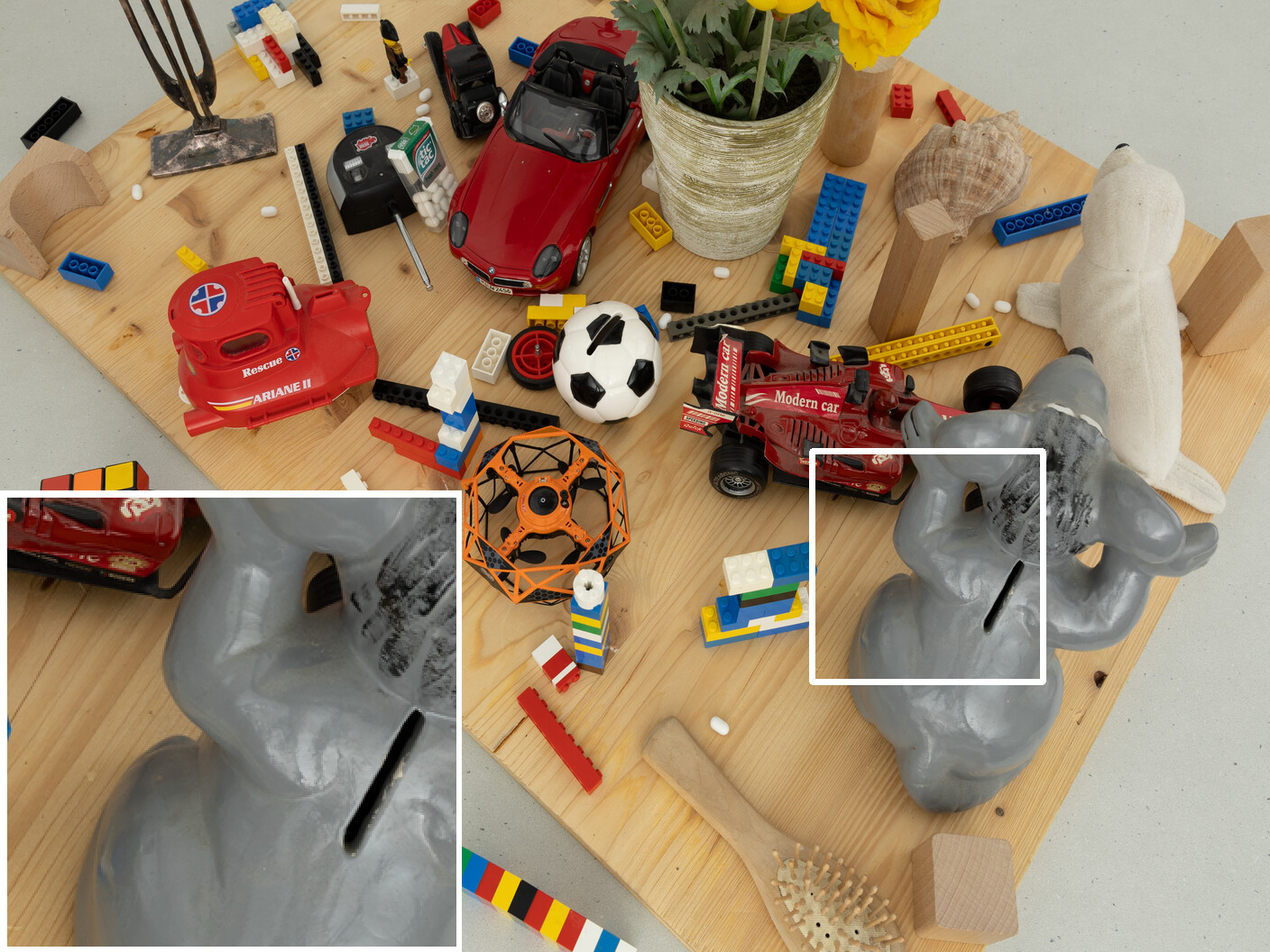} \\
\includegraphics[width=\widthcomp\linewidth]{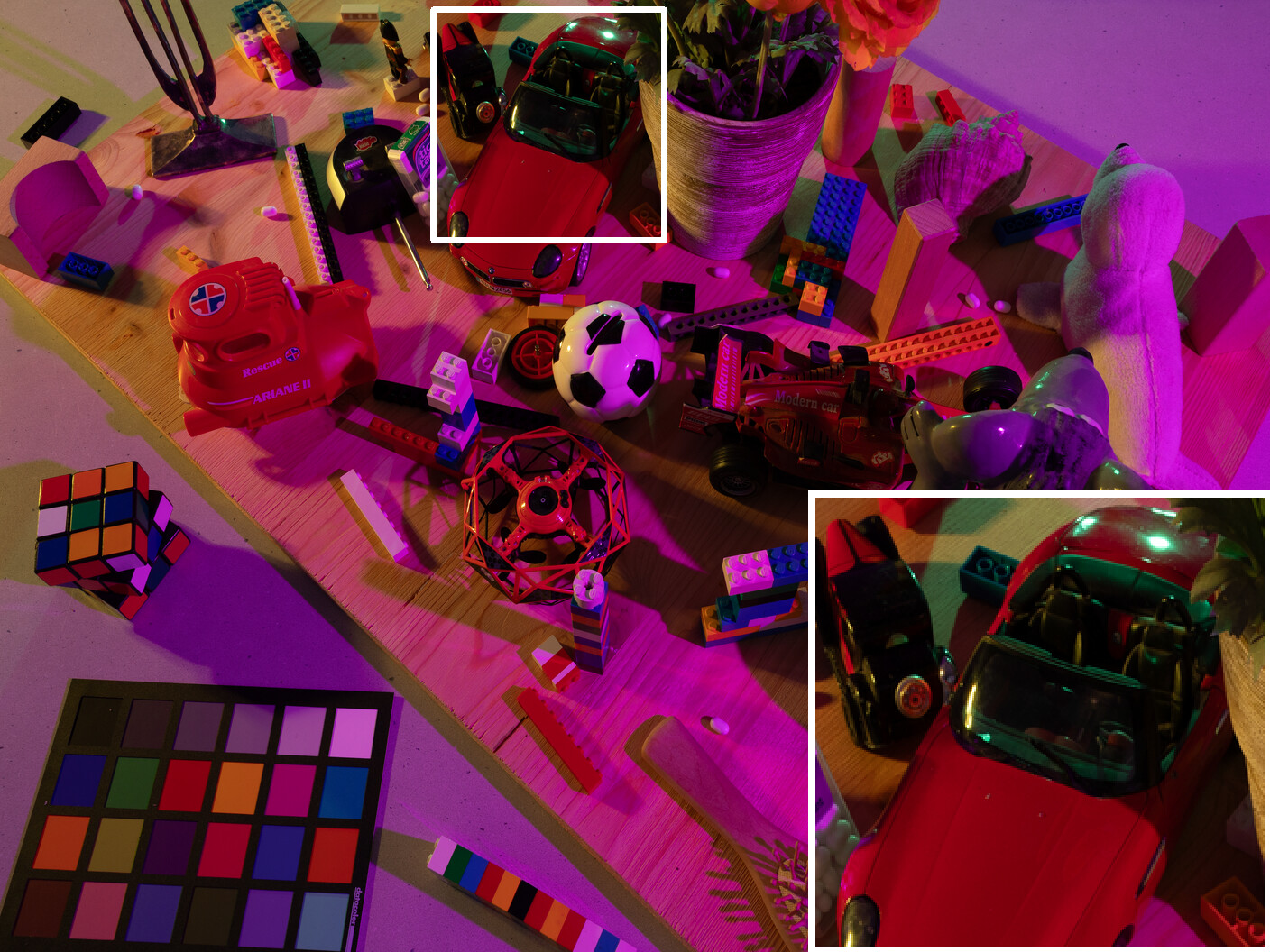} &
\includegraphics[width=\widthcomp\linewidth]{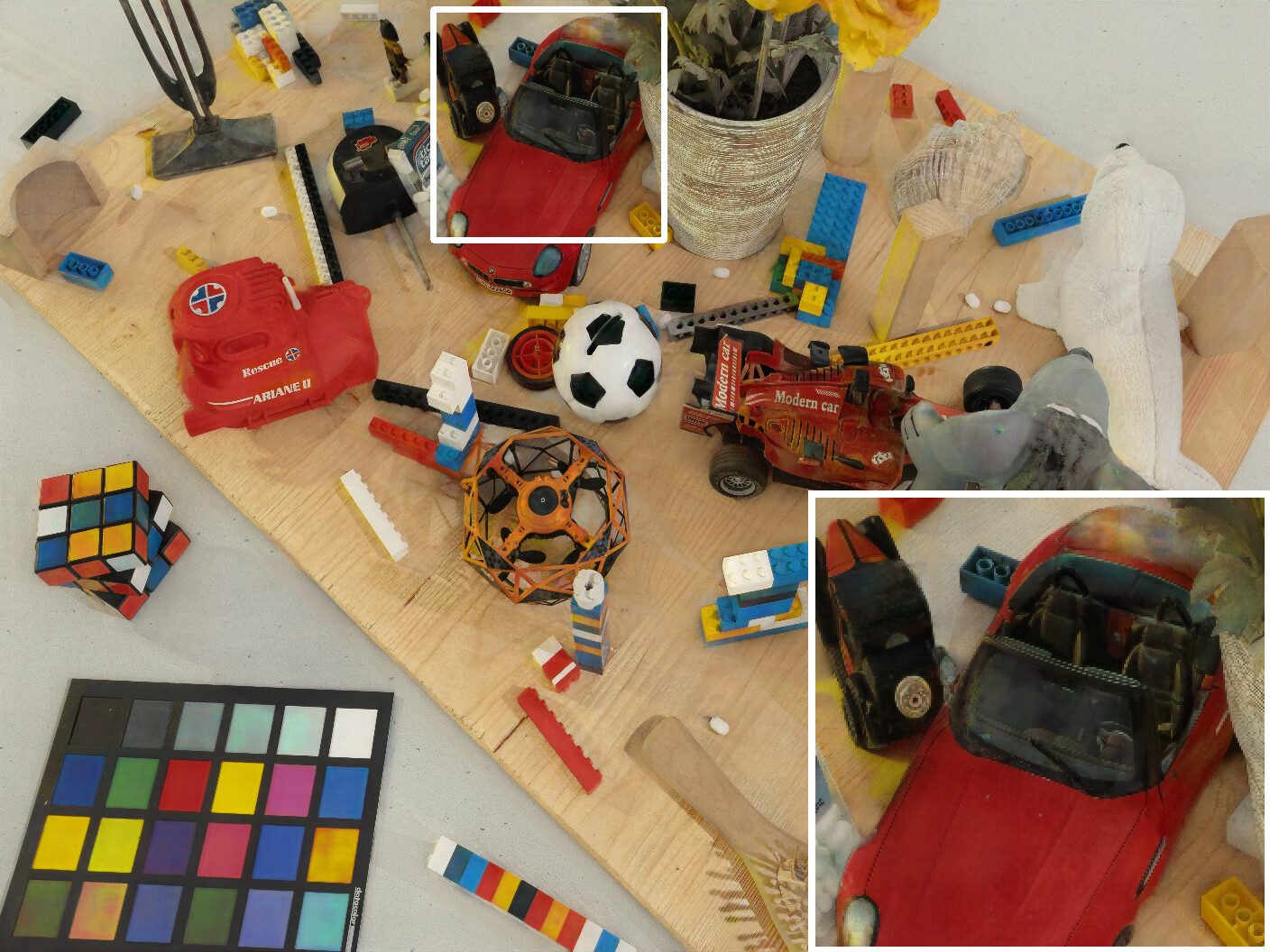} &
\includegraphics[width=\widthcomp\linewidth]{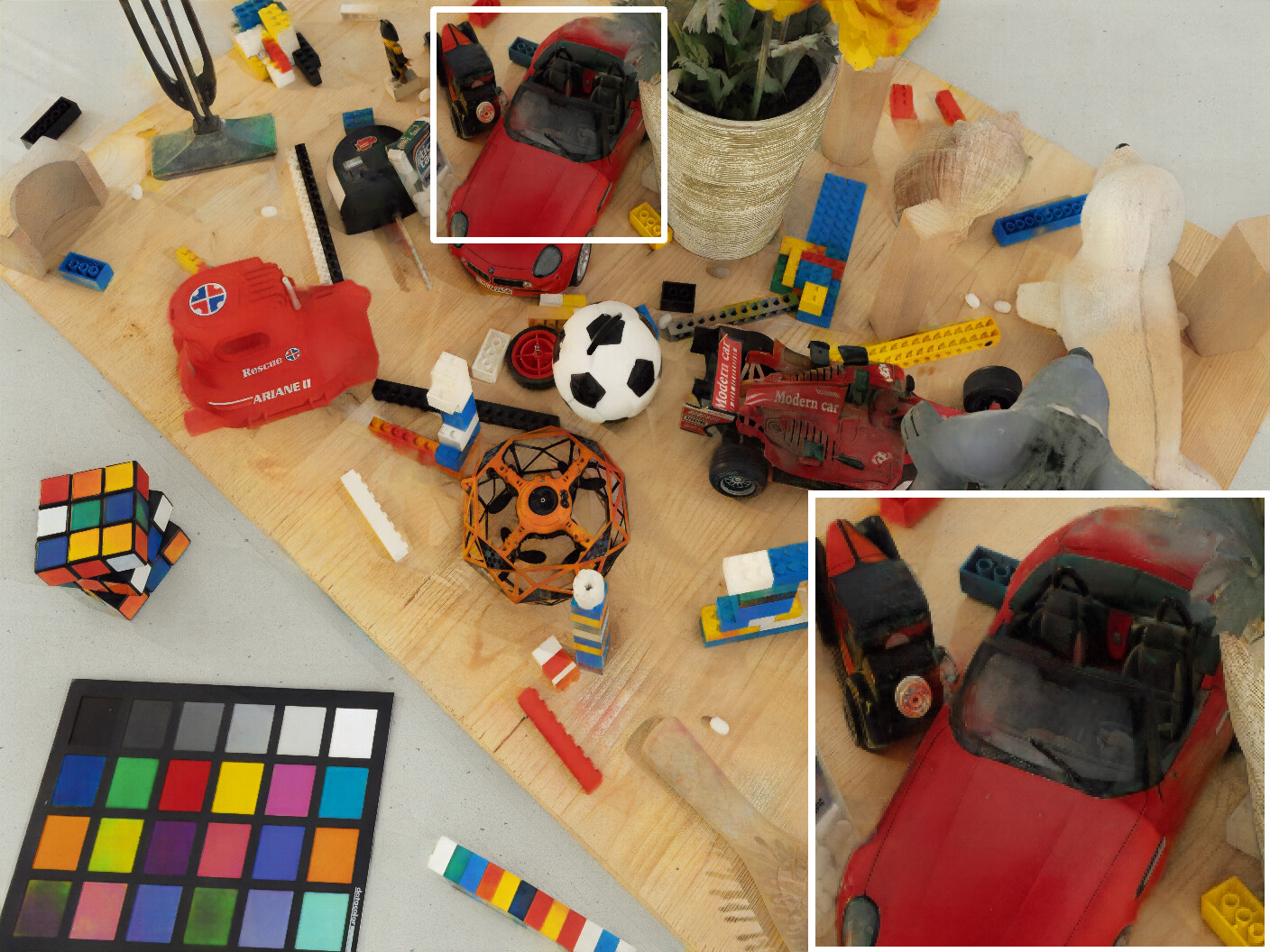} &
\includegraphics[width=\widthcomp\linewidth]{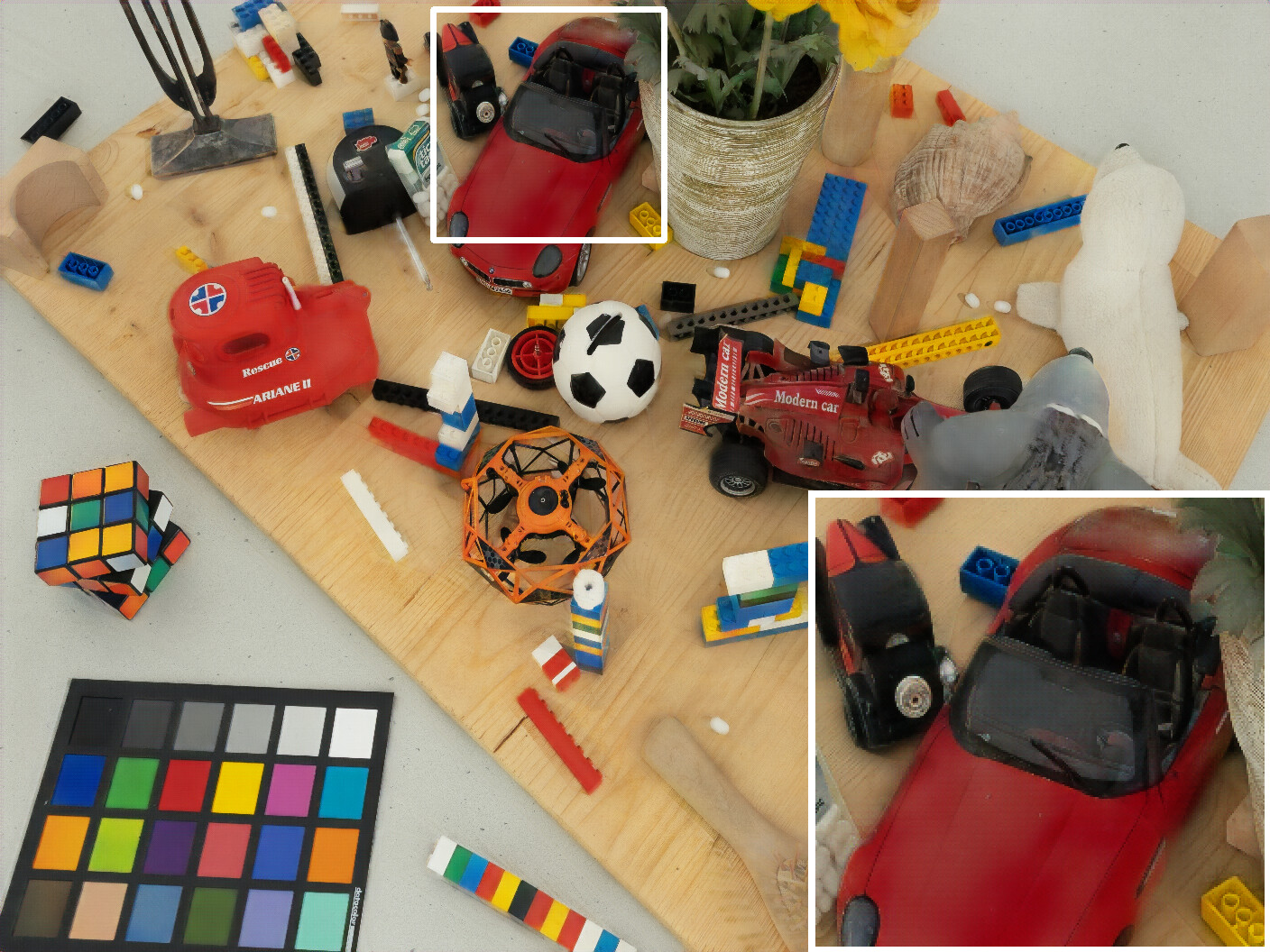} &
\includegraphics[width=\widthcomp\linewidth]{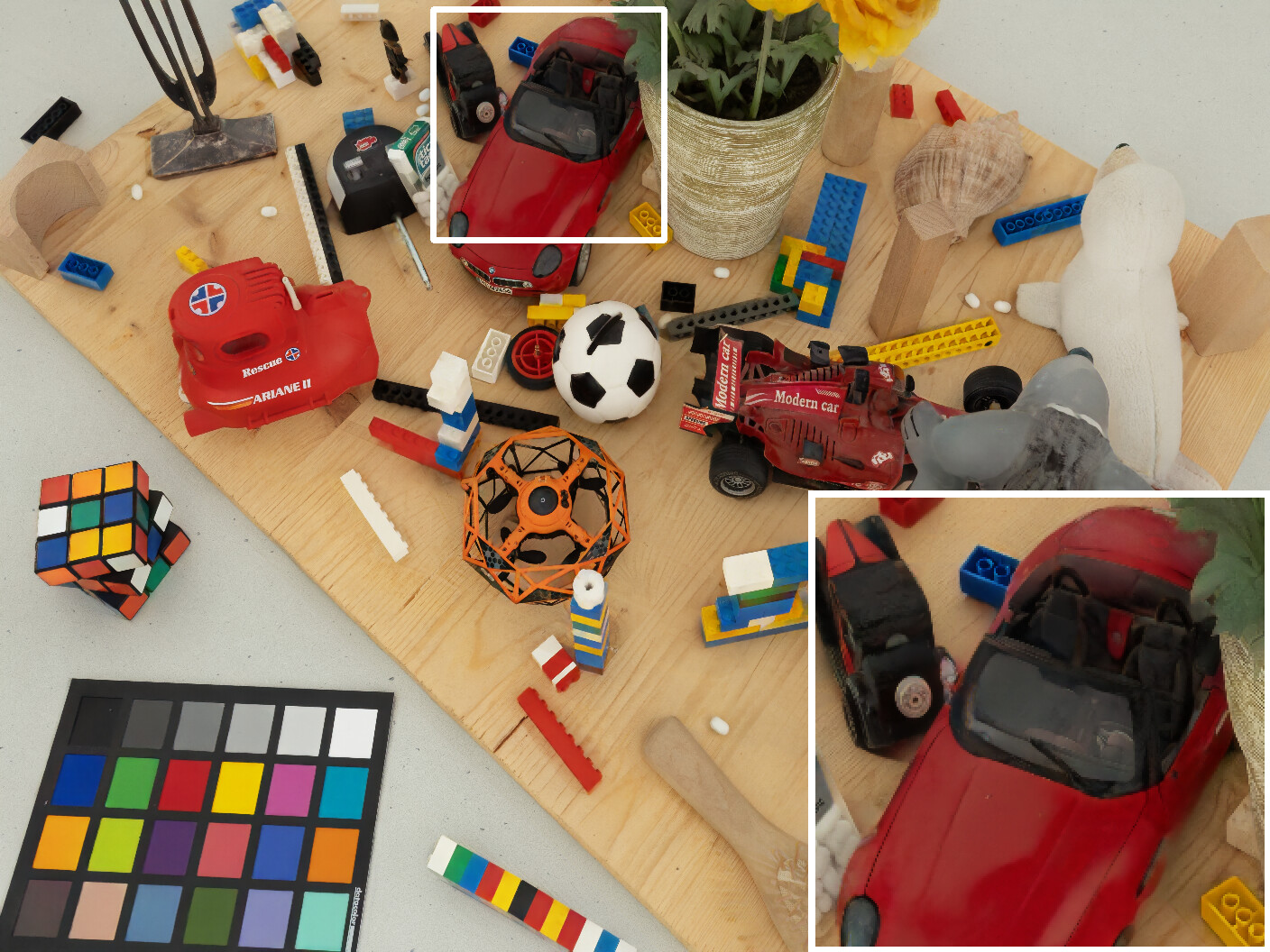} & 
\includegraphics[width=\widthcomp\linewidth]{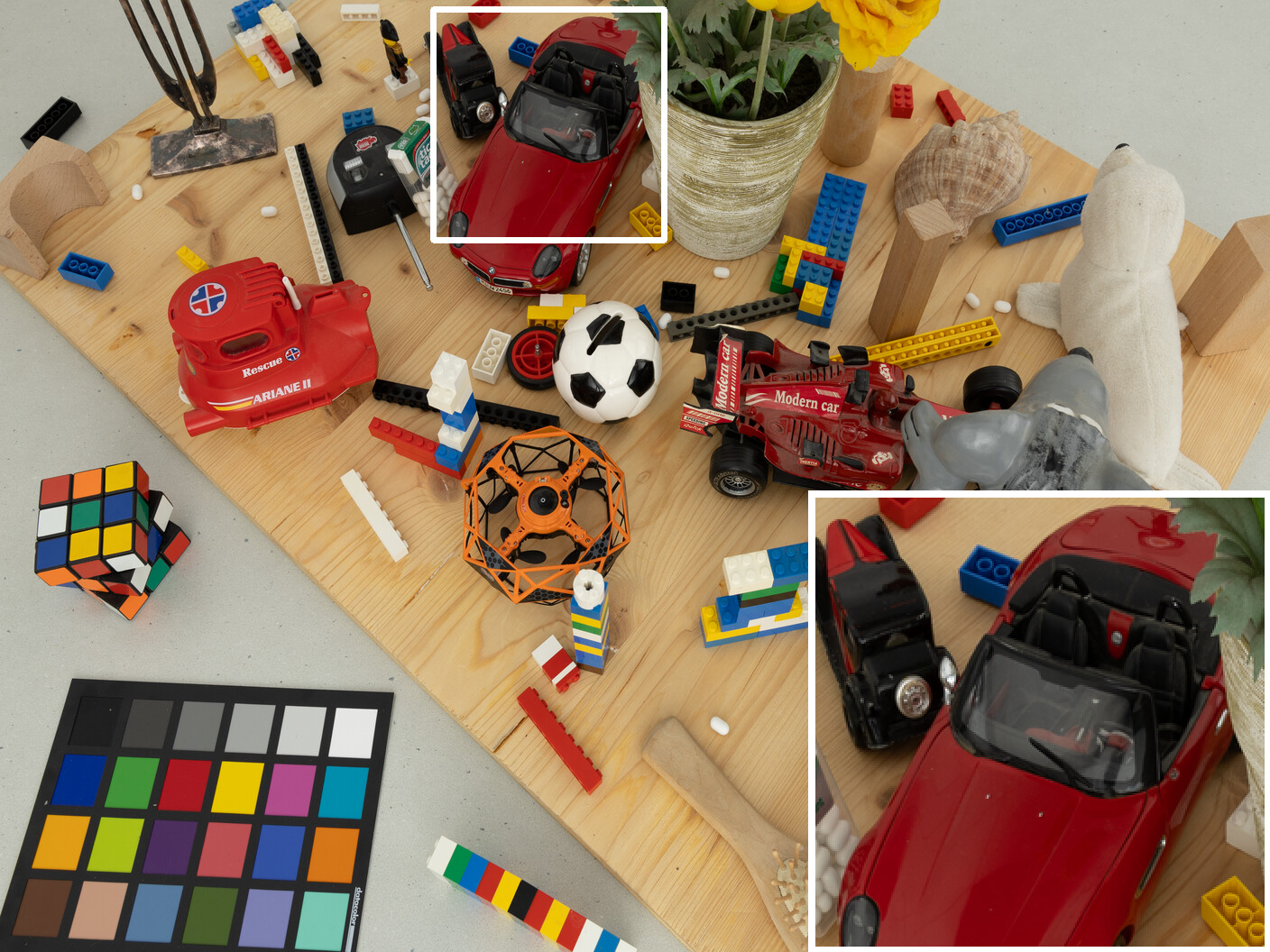} \\
\includegraphics[width=\widthcomp\linewidth]{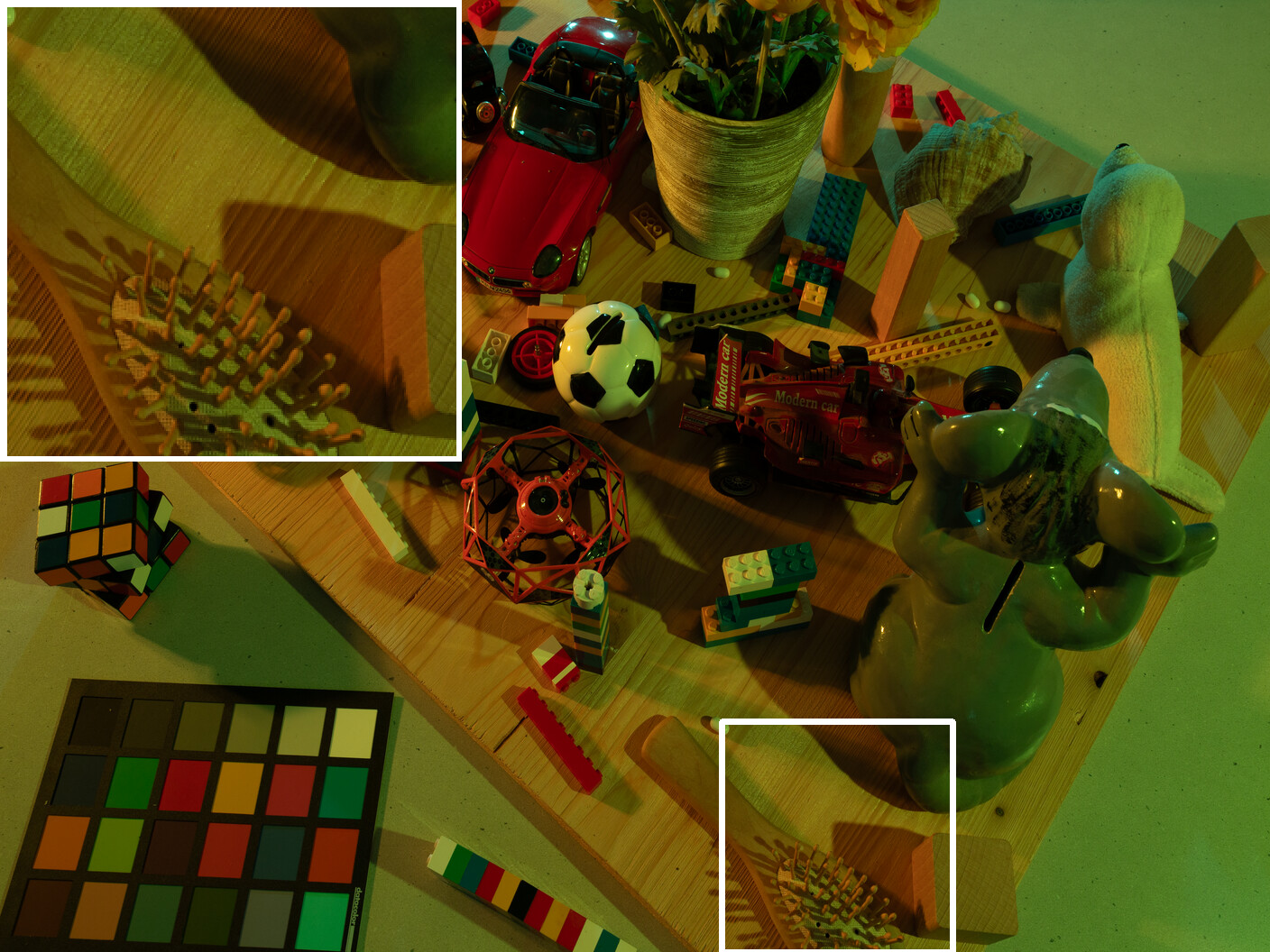} &
\includegraphics[width=\widthcomp\linewidth]{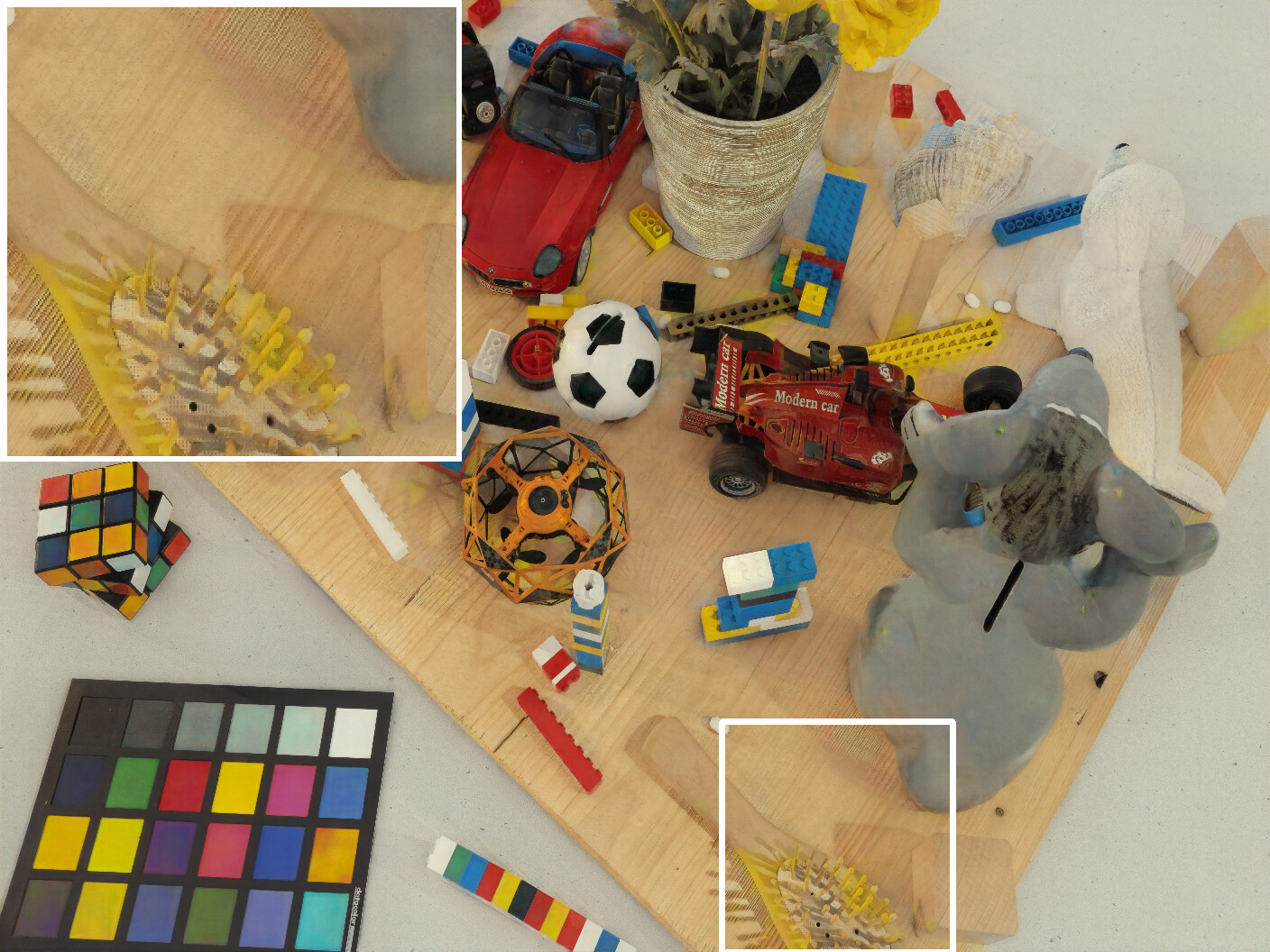} &
\includegraphics[width=\widthcomp\linewidth]{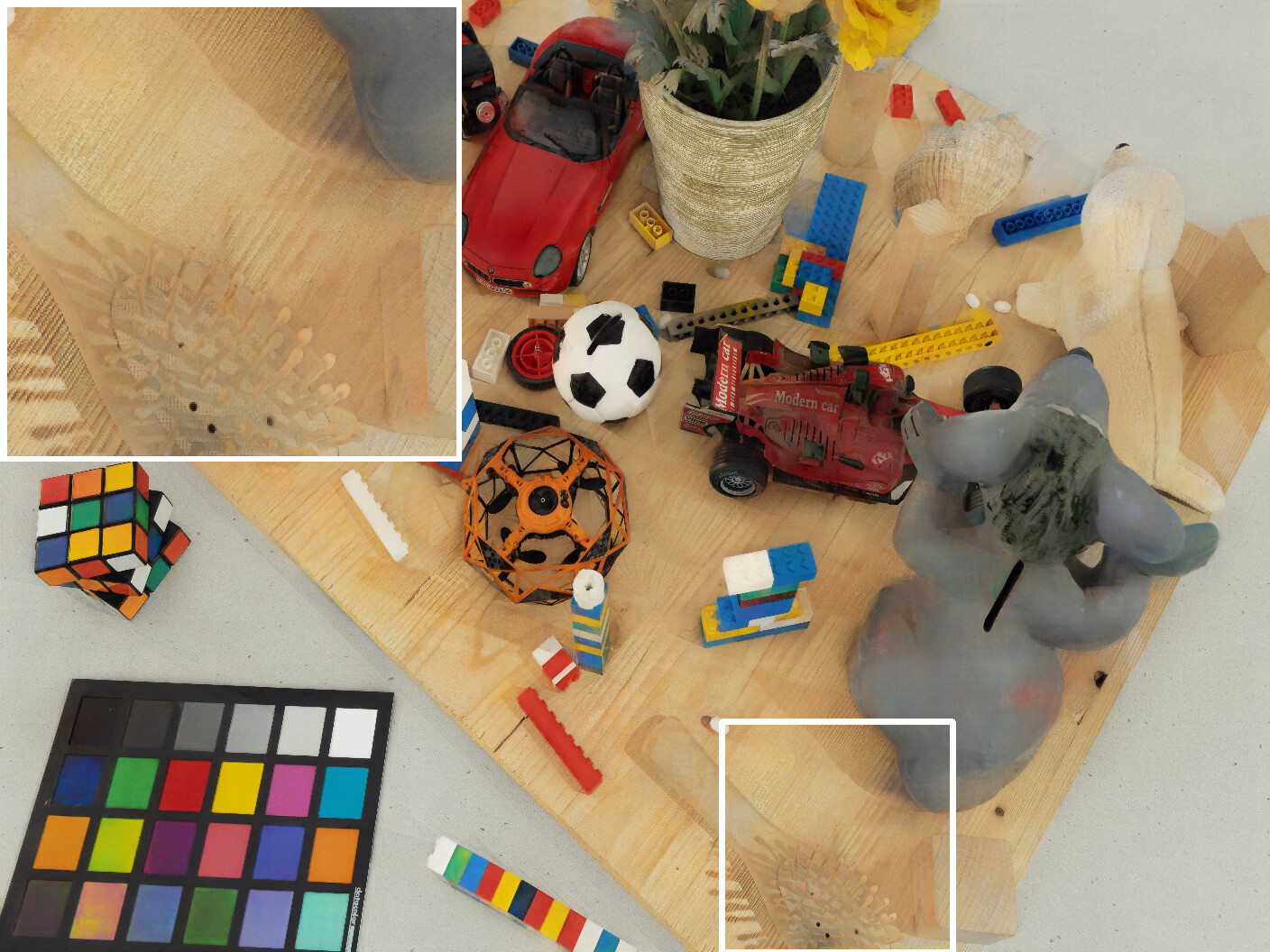} &
\includegraphics[width=\widthcomp\linewidth]{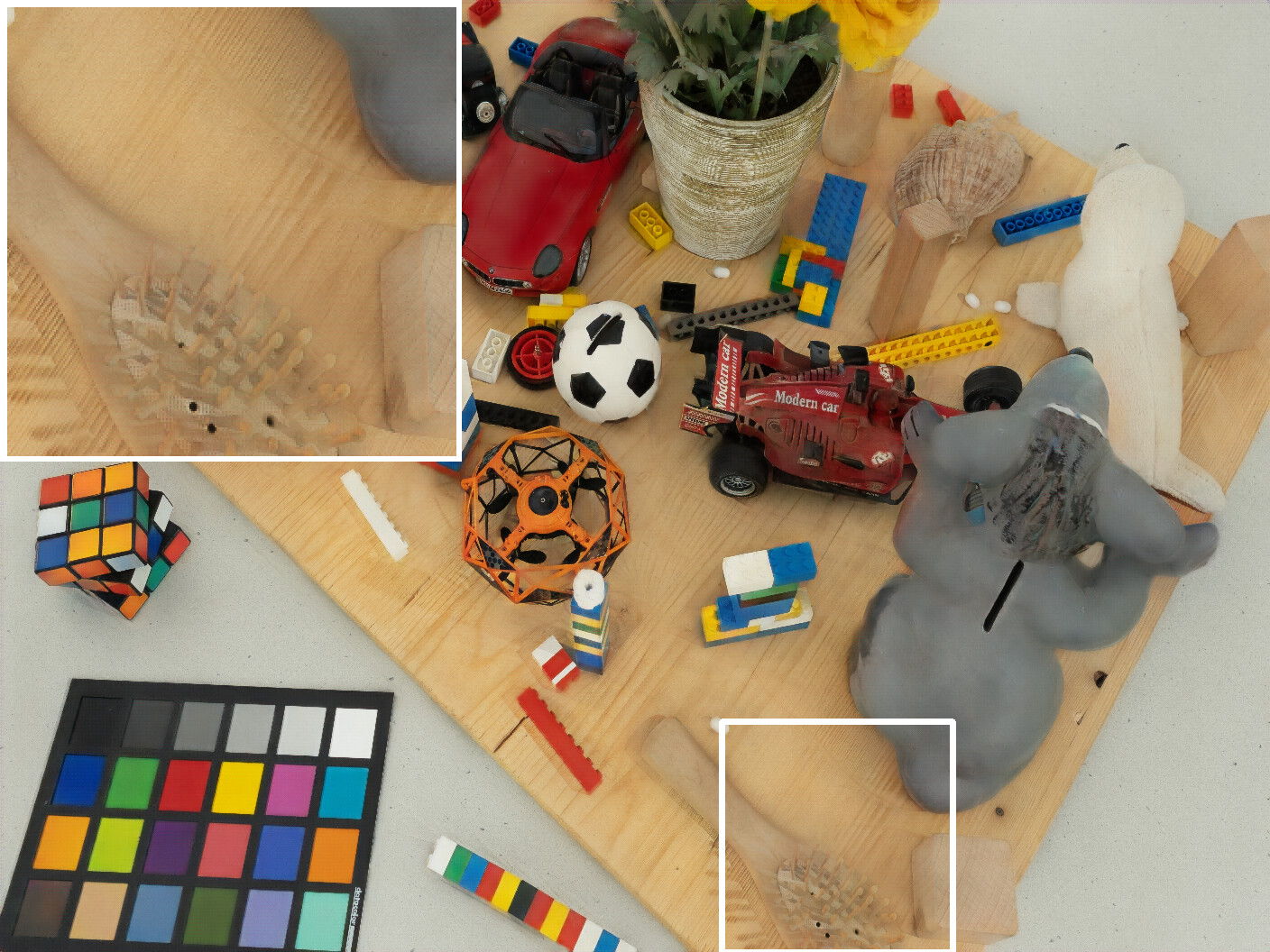} &
\includegraphics[width=\widthcomp\linewidth]{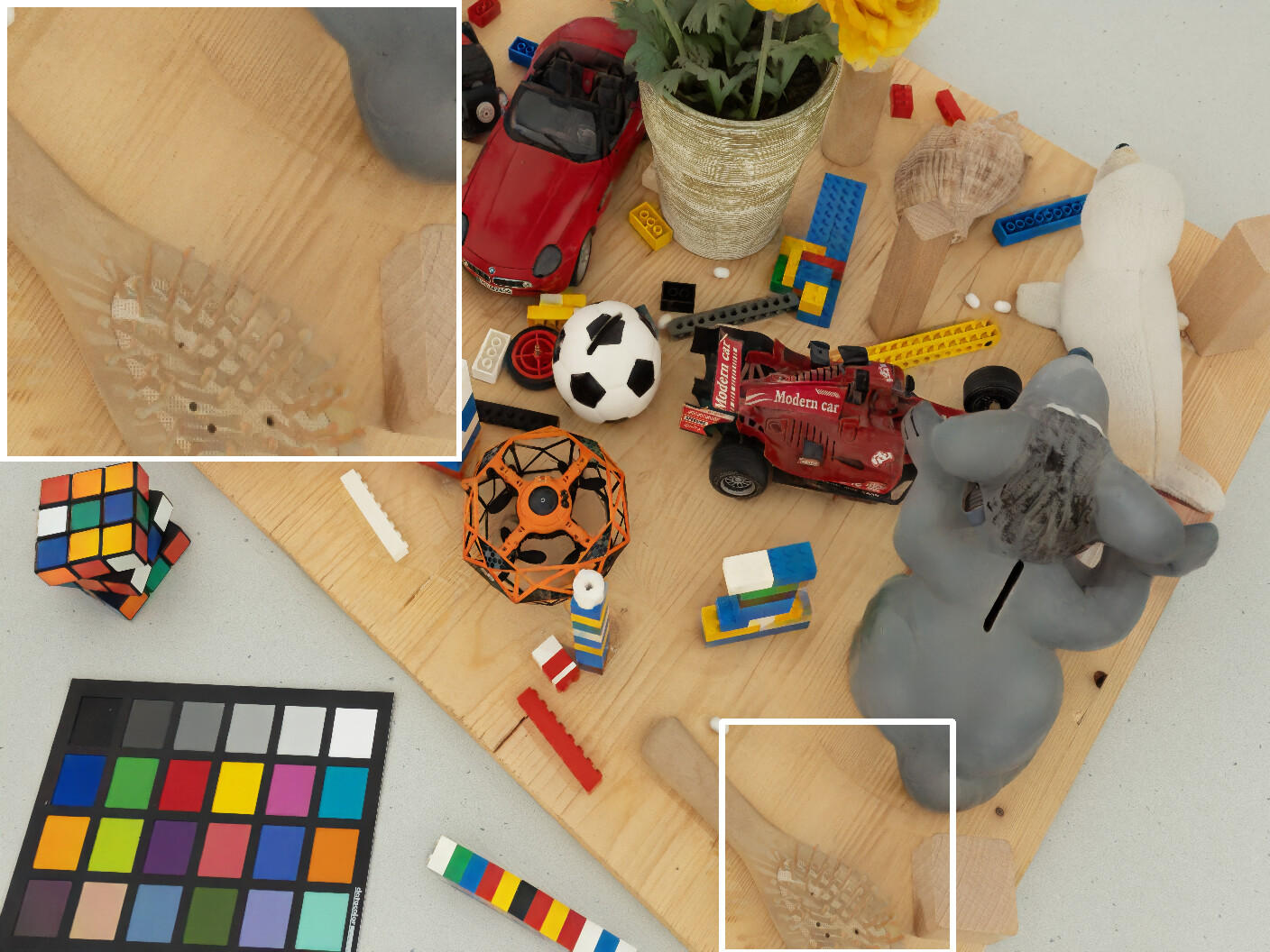} & 
\includegraphics[width=\widthcomp\linewidth]{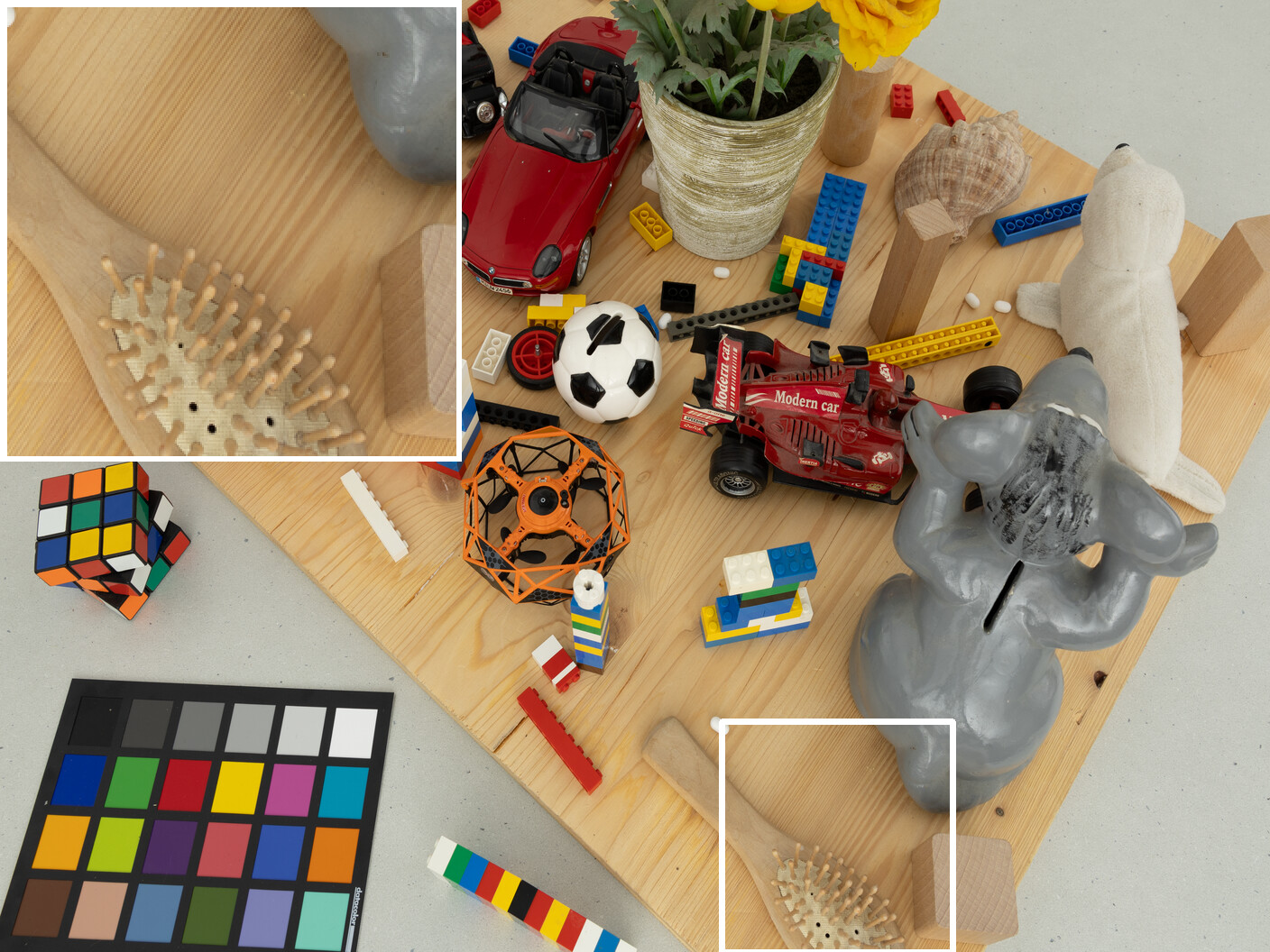} 

    \end{tabular}
    \vspace{-4mm}
    \caption{\textbf{Qualitative comparison} on scene \#1 from test split of the \textbf{CL3AN} dataset. Best viewed in the electronic version. }
    \label{fig:qualitative-cle3n}
    \vspace{-2mm}
\end{figure*}

\begin{table*}
\centering
\setlength{\tabcolsep}{3pt}

\vspace{-1mm}
\resizebox{\linewidth}{!}
{
\begin{tabular}{@{}c|cccc|ccc||ccc@{}}
\toprule
Method & \multicolumn{4}{c|}{General information} & \multicolumn{3}{c||}{CL3AN \emph{(ours)}} & \multicolumn{3}{c}{AMBIENT6K \cite{vasluianu2024towards}}\\
   name  & Type   &  Restoration Task   & Prior  & MACs (G.) & \text{PSNR}$\uparrow$ & \text{SSIM}$\uparrow$ & \text{LPIPS}$\downarrow$ & \text{PSNR}$\uparrow$ & \text{SSIM}$\uparrow$ & \text{LPIPS}$\downarrow$\\
\midrule
unprocessed               & -   & -  & -   & -  & 10.837 & 0.447 & 0.518 & 13.403                & 0.652                 & 0.250   \\
\midrule
NAFNet\cite{chen2022simple}   & Conv.  & Multi-task IR & RGB & 15.92  & 19.476 & 0.709 & 0.249     & 20.580                & 0.808                 & 0.142    \\
MPRNet \cite{zamir2021multi} & Conv. & Multi-task IR & RGB  & 37.21   & 18.453  & 0.688 & 0.291   & 20.947                 & 0.820                 & 0.129   \\

SFNet \cite{cui2023selective}  & Transf. & Multi-task IR & RGB + Freq.  & 31.27   & 18.382 & 0.686 & 0.291    & 20.519                & 0.812                 & 0.141    \\
SwinIR \cite{liang2021swinir} & Transf. & Multi-task IR  & RGB  & 37.81  & 16.386 & 0.643 & 0.372     & 20.528                & 0.817                 & 0.131    \\
Uformer \cite{wang2022uformer} & Transf. & Multi-task IR & RGB  & 19.33    & 17.508 & 0.655 & 0.313   & 20.776                & 0.818                 & 0.131    \\
Restormer \cite{zamir2022restormer}  & Transf. & Multi-task IR & RGB & 35.31 & 18.560 & 0.691 & 0.278    & 21.141               & 0.817                 & 0.132      \\
HINet \cite{jing2021hinet}  & Conv. & Multi-task IR & RGB & 42.68   & 19.388 & 0.707 & 0.248    & 20.856                & \underline{0.821}                 & 0.129      \\
IFBlend \cite{vasluianu2024towards}   & Conv & ALN  & RGB + Freq.  & 26.01 & 20.370 & 0.720 & 0.228       & 21.443                & 0.819        & 0.128 \\
MAMBAIR \cite{guo2025mambair} & Transf. + SSM & Multi-task IR & RGB & 34.32 & 18.970 & 0.704 & 0.254 & - & - & -\\
GRL \cite{li2023grl} & Transf. & Multi-task IR & RGB & 2.16 &  18.089 & 0.672 & 0.308 & - & - & -\\
Retinexformer \cite{cai2023retinexformer} & Transf. & LLIE & RGB & 4.86 & 18.649 & 0.683 & 0.281 & - & - & -\\
\midrule
RLN$^2$-S (\emph{ours}) & Conv & ALN  & RGB  & 3.95 & 19.992 & 0.718 & 0.236 & 21.181 & 0.815 & 0.131\\
RLN$^2$-Sf (\emph{ours}) & Conv & ALN  & RGB + Freq. & 4.24 & 20.128 & \underline{0.730} & 0.223 &  21.333 & 0.819 & 0.128\\
\midrule

RLN$^2$-L (\emph{ours}) & Conv & ALN  & RGB & 22.45 & \underline{20.383} & 0.723 & \underline{0.222} & \underline{21.553} & \underline{0.821} & \underline{0.123}\\
RLN$^2$-Lf (\emph{ours}) & Conv & ALN  & RGB + Freq. & 22.72 & \textbf{20.523} & \textbf{0.746} & \textbf{0.208} & \textbf{21.712} & \textbf{0.825} & \textbf{0.120} \\
\bottomrule
\end{tabular}%
}
\vspace{-2mm}
\caption{Various RLN$^2$ variants versus current SOTA image restoration models, on the proposed \textbf{CL3AN} dataset test split (1920 $\times$ 1440 px.), and AMBIENT6K benchmark \cite{vasluianu2024towards} (1280 $\times$ 960 px. default size). The \textbf{best} and the \underline{second best} performing models are emphasized.}
\vspace{-5mm}
\label{tab:quanti-test}
\end{table*}

\begin{figure*}[h!]
    \centering
    \setlength{\tabcolsep}{1pt}
    \renewcommand{\arraystretch}{0.7}
    \def\widthcomp{0.2}
    \begin{tabular}{ccccc}
         Input     &  Restormer \cite{zamir2022restormer}   & IFBlend \cite{vasluianu2024towards} & RLN$^2$-Lf \emph{(Ours)}  & Ground Truth                                                                   \\
\includegraphics[width=\widthcomp\linewidth]{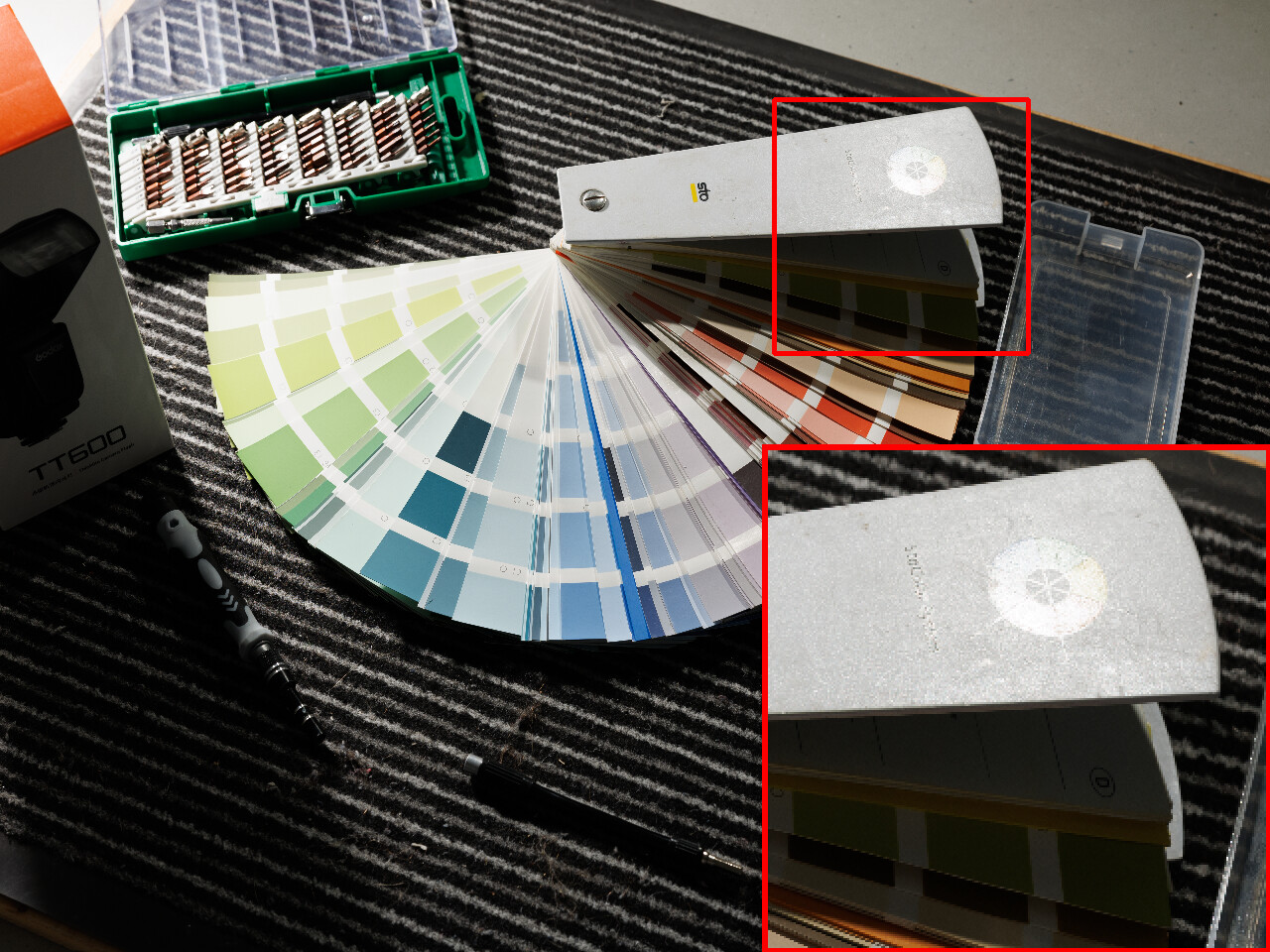} & \includegraphics[width=\widthcomp\linewidth]{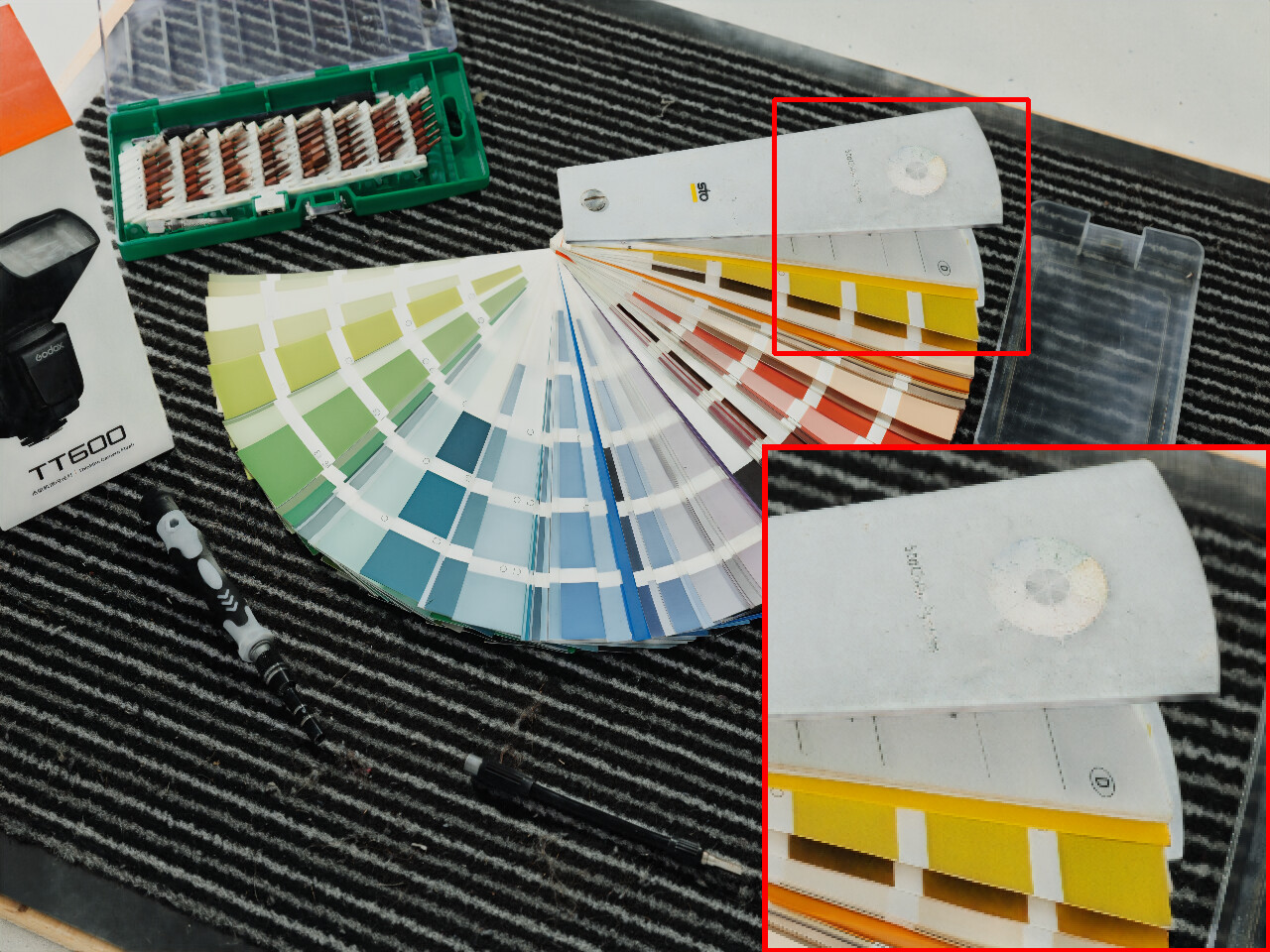} & \includegraphics[width=\widthcomp\linewidth]{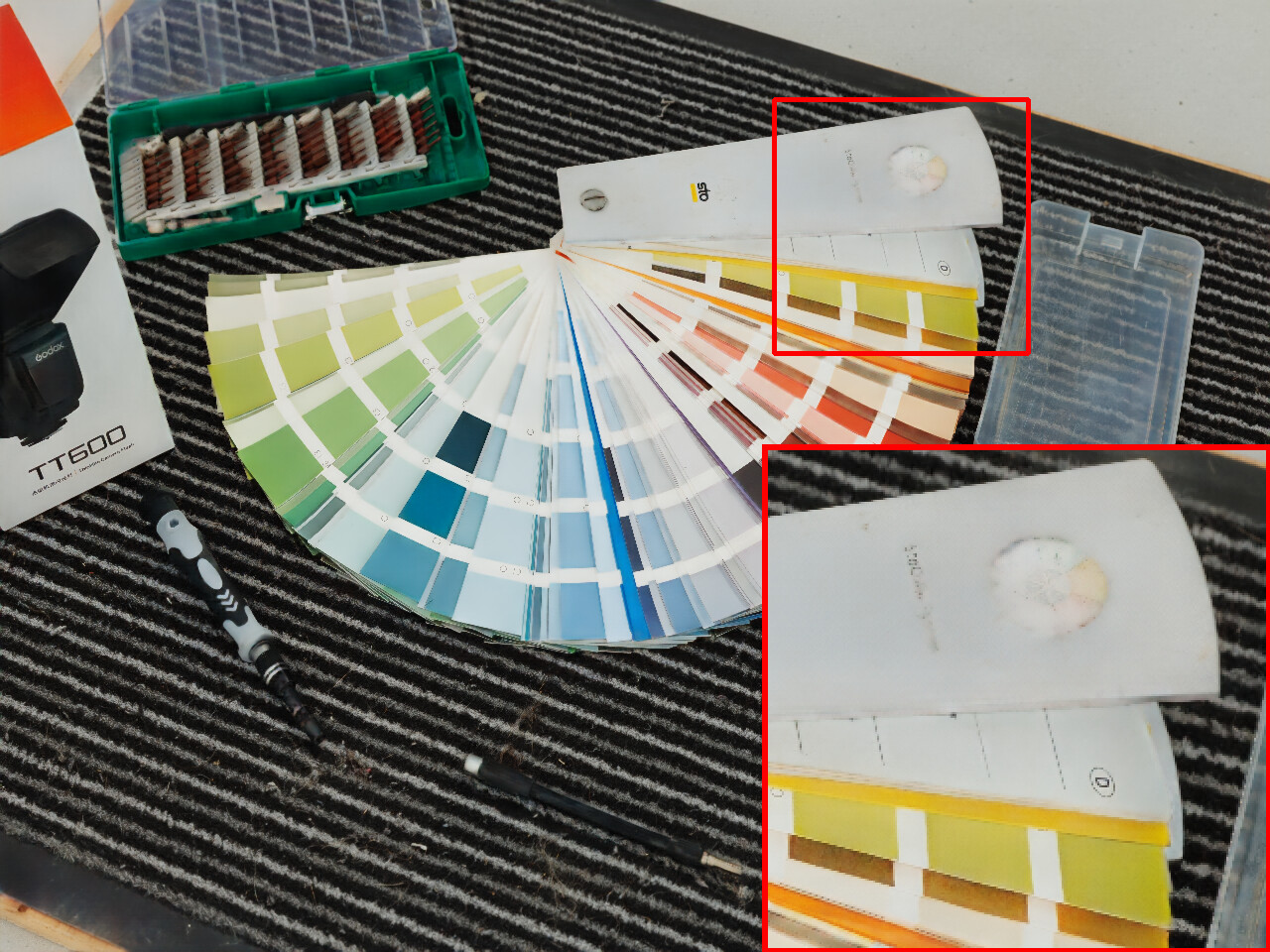} & \includegraphics[width=\widthcomp\linewidth]{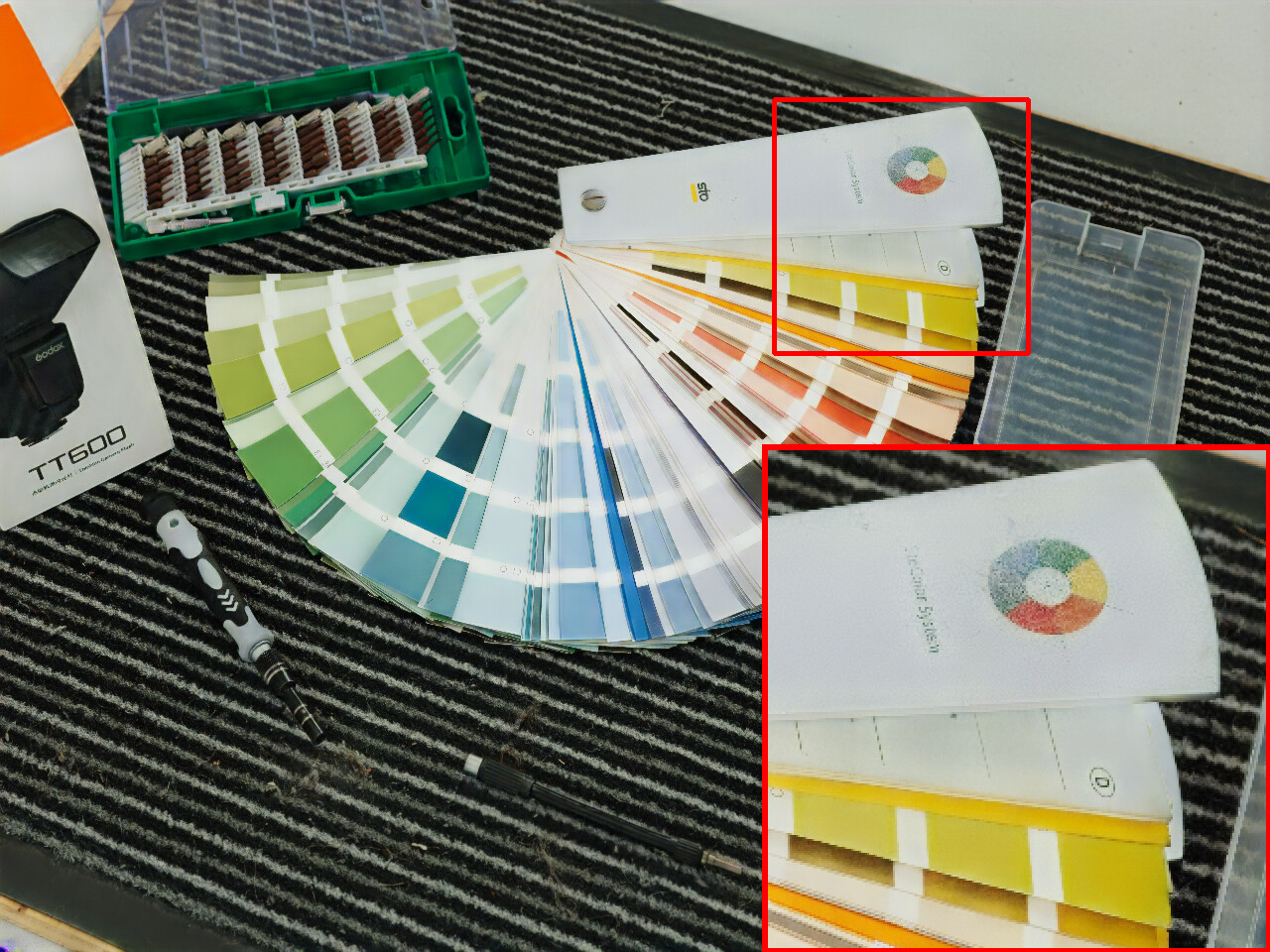} & \includegraphics[width=\widthcomp\linewidth]{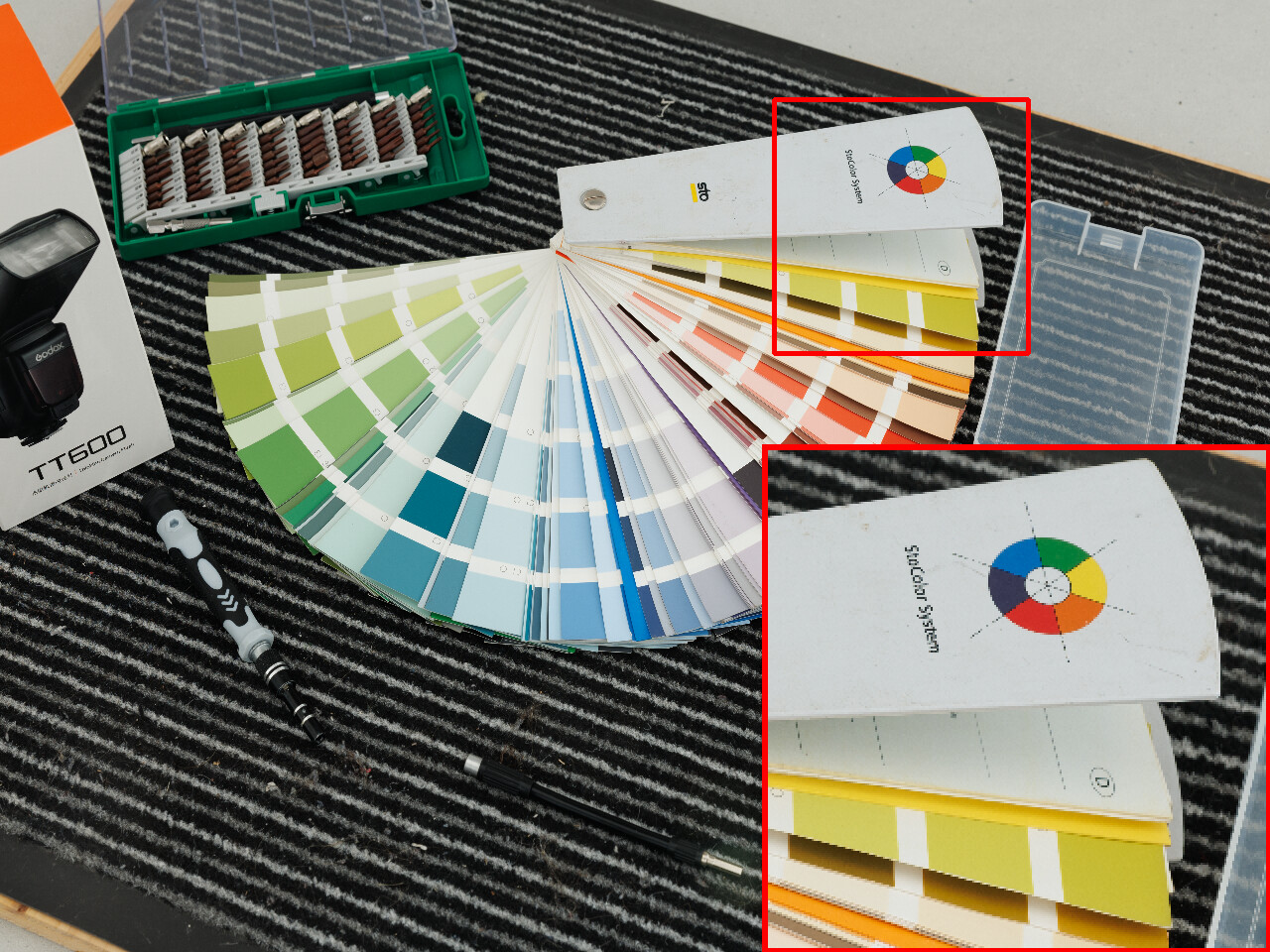} \\
\includegraphics[width=\widthcomp\linewidth]{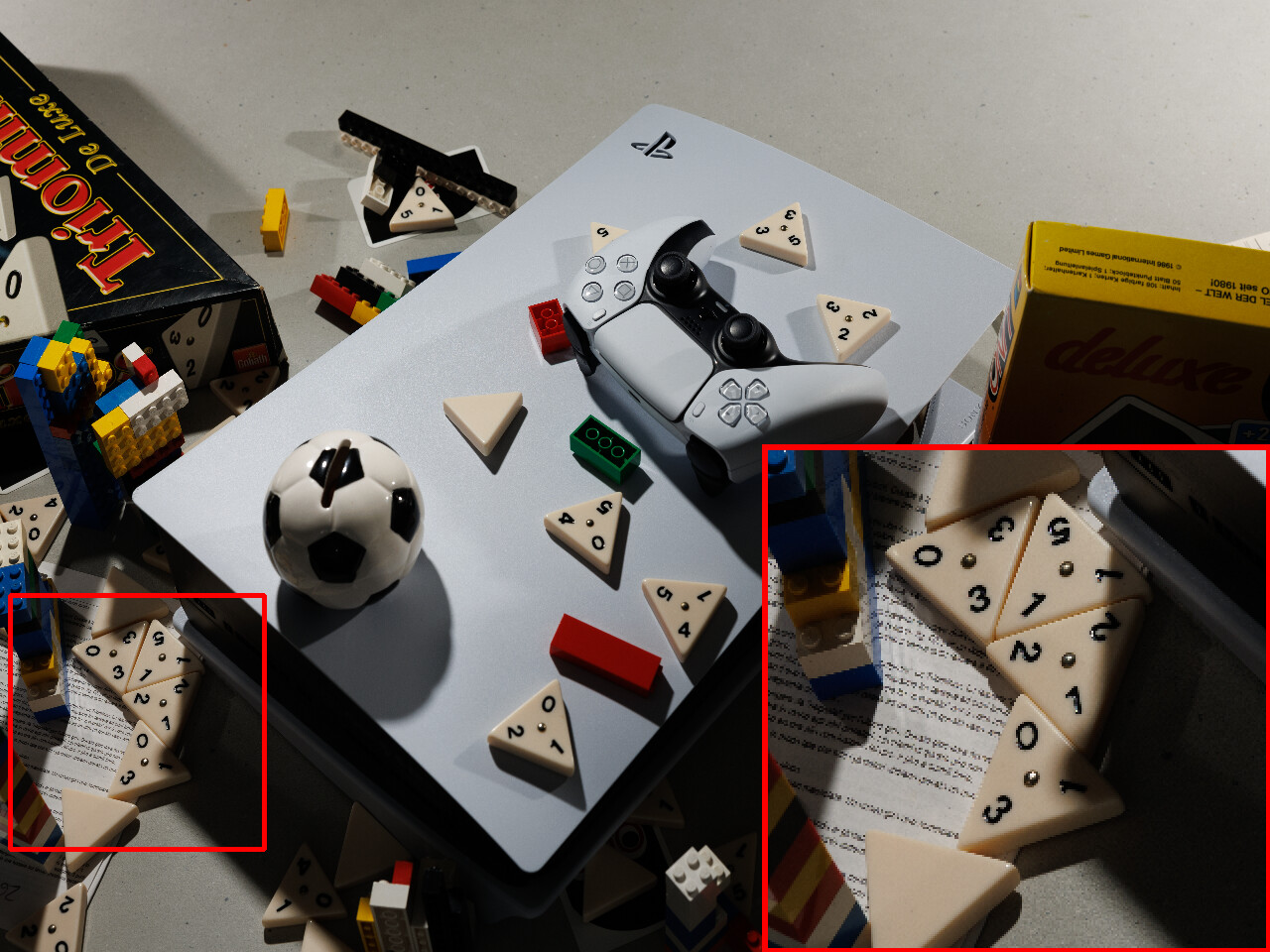} & \includegraphics[width=\widthcomp\linewidth]{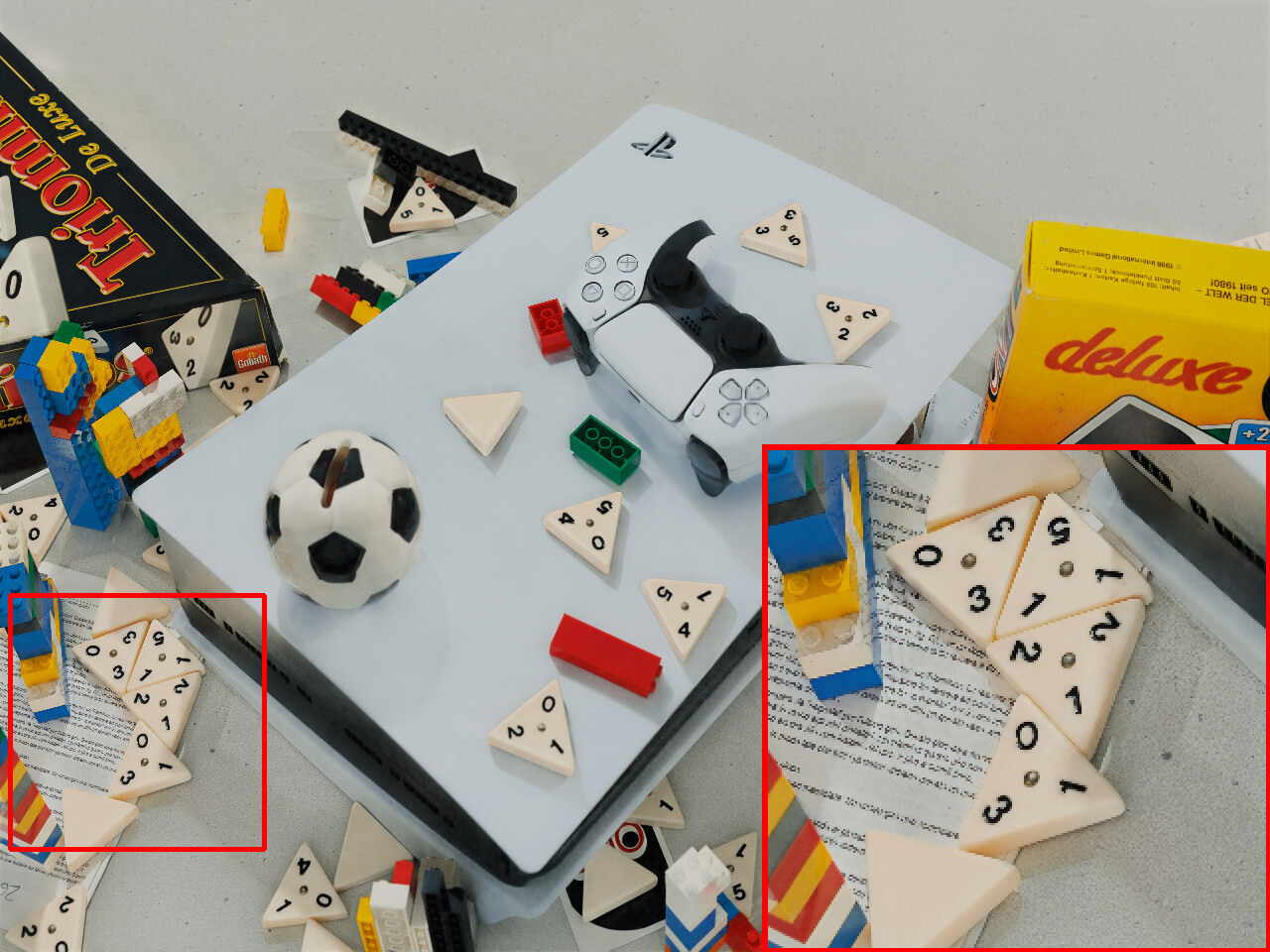} & \includegraphics[width=\widthcomp\linewidth]{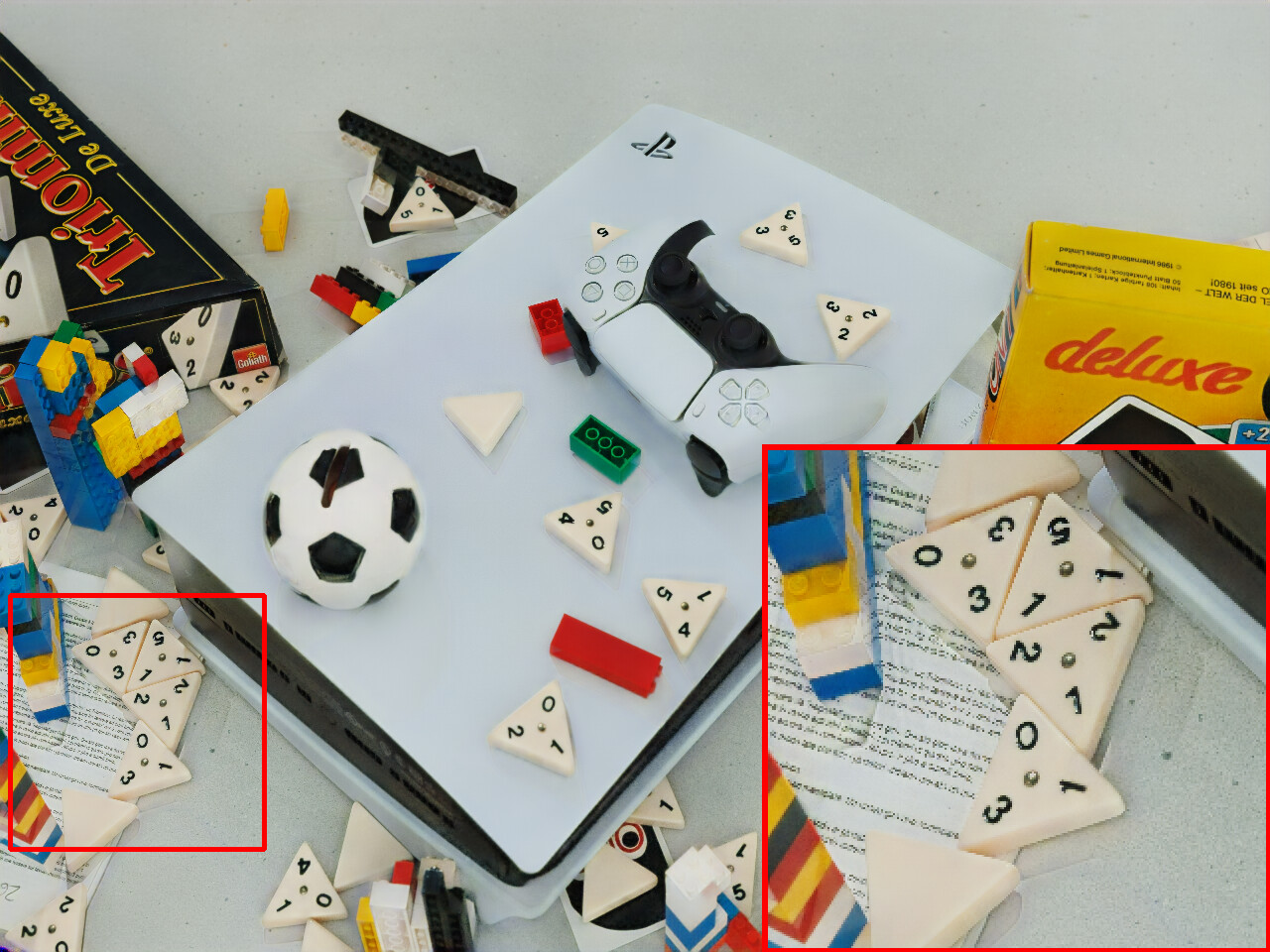} & \includegraphics[width=\widthcomp\linewidth]{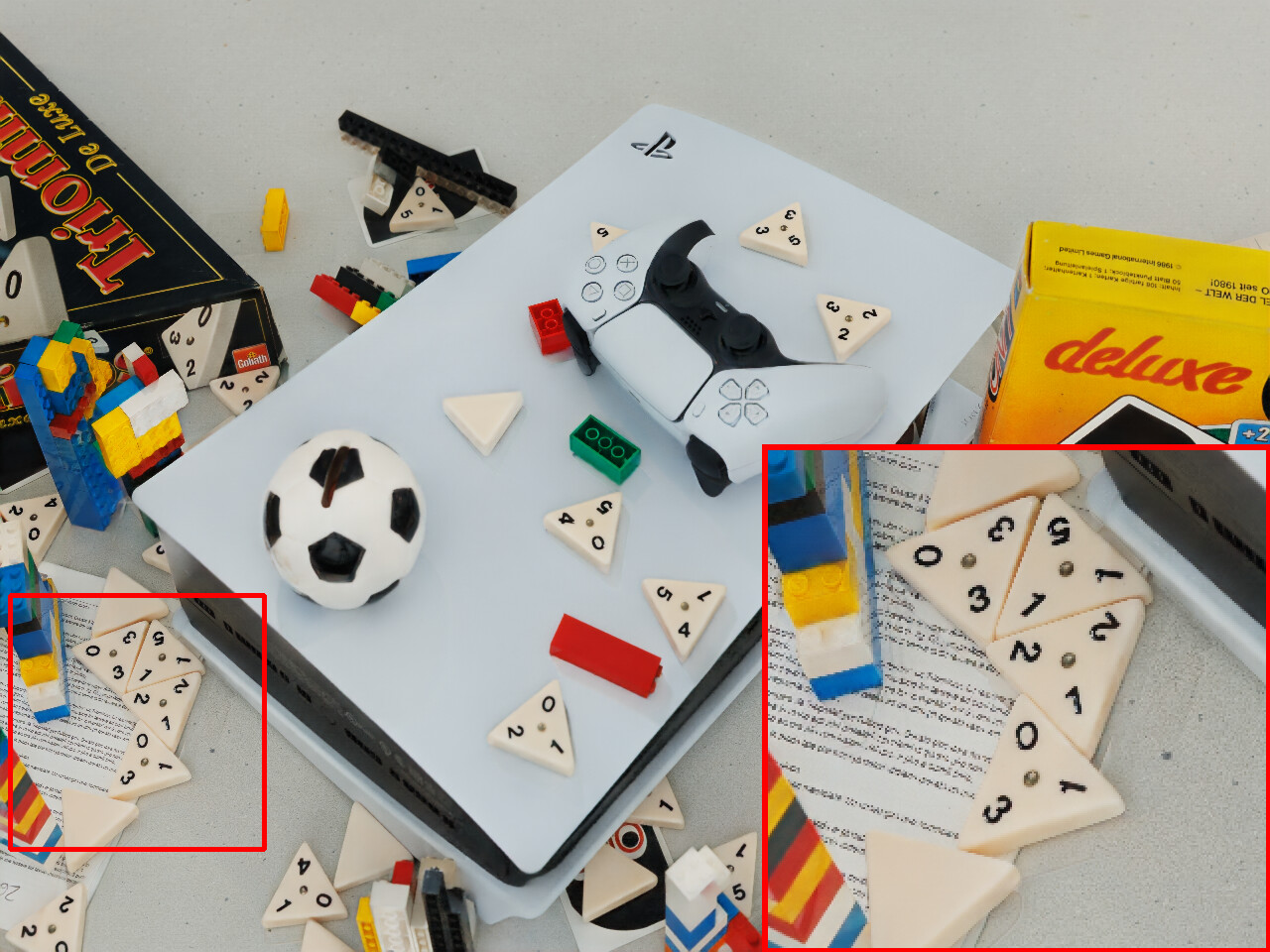} & \includegraphics[width=\widthcomp\linewidth]{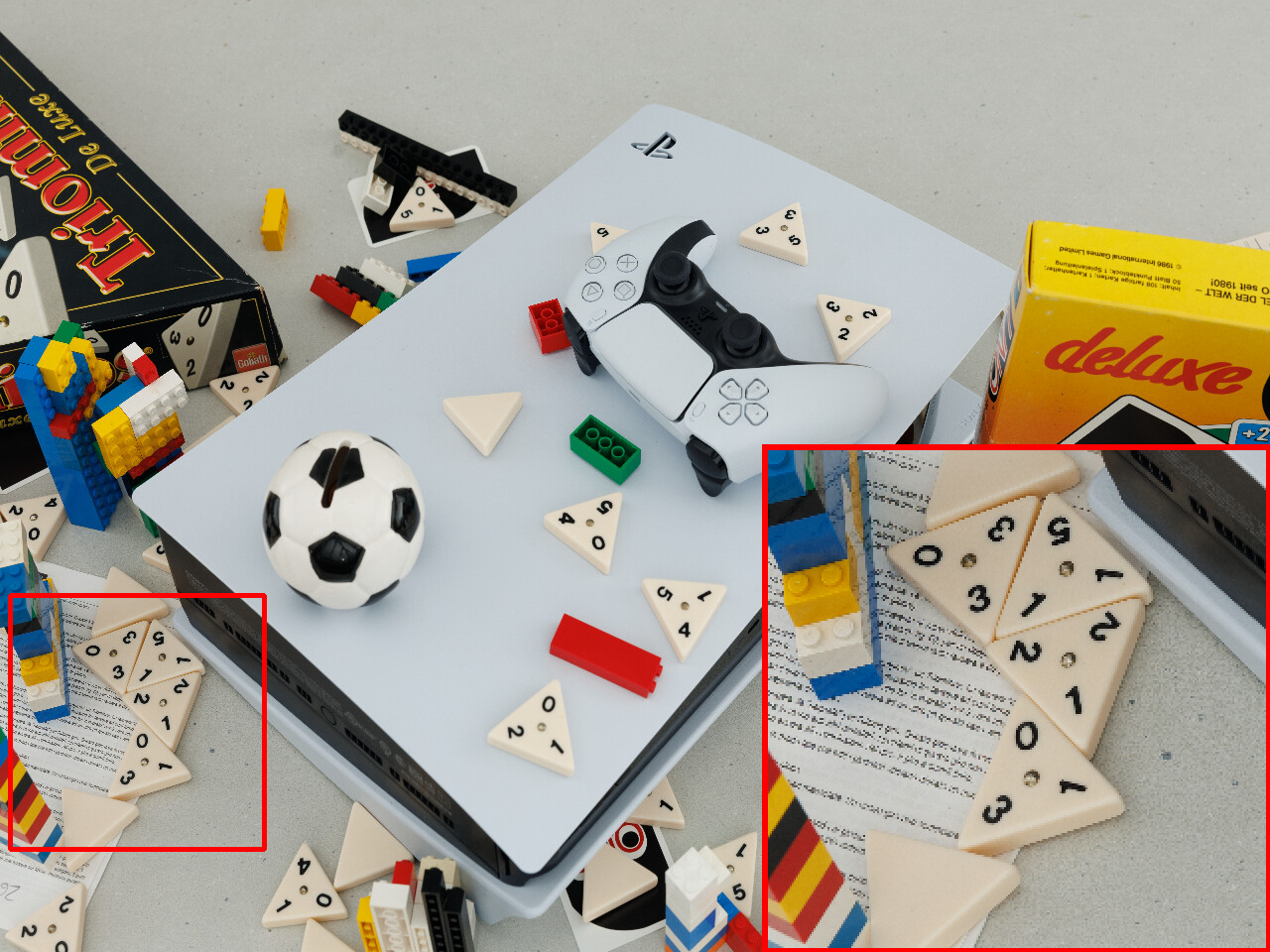} \\
    \end{tabular}
    \vspace{-4mm}
    \caption{Qualitative comparison challenging samples from the AMBIENT6K testing set \cite{vasluianu2024towards}. Our RLN$^2$ produces improved quality ambient normalized images, with sharper images, and a lower amount of visible artifacts.}
    \label{fig:qualitative-ambient}
    \vspace{-5mm}
\end{figure*}

\noindent\textbf{Experimental Setup:} Additionally to the introduced \textbf{CL3AN} dataset, we perform experiments on the AMBIENT6K image database \cite{vasluianu2024towards}, which consists of 6000 images representing scenes under the effect of up to three white-aligned LED directional lights (see \cref{fig:dataset_fig}). 

As compared metrics, we report restoration fidelity through Peak Signal to Noise Ratio (PSNR), and Structure Similarity Index Measure (SSIM) \cite{wang2004image}. To quantify the perceptual quality of the restored images, we deploy the AlexNet-based Learned Perceptual Image Patch Similarity score (LPIPS) \cite{zhang2018perceptual}. 

In terms of compared methods, we chose various types of solutions proposed in the image restoration field, which became seminal work after their introduction. For an extensive benchmark, we combine refined CNN solutions such as MPRNet \cite{zamir2021multi} and HINet 
\cite{jing2021hinet}, with transformer-based architectures like SwinIR \cite{liang2021swinir}, NAFNet \cite{chen2022simple}, Uformer \cite{wang2022uformer}, Restormer \cite{zamir2022restormer}, and GRL \cite{li2023grl}. Additionally, we consider solutions integrating state-space-models (\eg MAMBAIR \cite{gu2023mamba}), or solutions optimizing joint Spatial-Frequency entropy such as SFNet \cite{cui2023selective} and IFBlend \cite{vasluianu2024towards}. Ultimately, we include Retinexformer \cite{cai2023retinexformer}, representing the class of solutions based on Retinex image decomposition. We train all the aforementioned methods using their official public implementations and settings. 

\noindent\textbf{Quantitative Results:} In \cref{tab:quanti-test}, we compare the aforementioned solutions with different variants of the proposed RLN$^2$. 
The different variants represent our proposed models at different complexity levels, such that when comparing to similar existing solutions, advantages become easily identifiable.
The computational complexity, as the Multiplications-Additions count (MAC), is reported for a patch of 128$\times$128 pixels, using \emph{ptflops} for consistency. 

The naming convention can be easily followed. 
Firstly, the difference between the RLN$^2$-S and the RLN$^2$-L is given by the complexity of the Wide Context Feature Extractor found in the feature fusion module, at the decoder stage, with RLN$^2$-S using a ConvNeXt-T configuration, while RLN$^2$-L benefits from a large  ConvNeXt-XL model \cite{liu2022convnet}.  
Secondly, we distinguish between RLN$^2$ variants that have/have not access to the dual image frequency domain representation. Thus RLN$^2$-S and RLN$^2$-L drop the frequency domain feature at the encoder, refinement stage, and decoder stage, using cross-attention for fusion, while RLN$^2$-Sf and RLN$^2$-Lf use all the available information.

All the compared RLN$^2$ variants are trained under the same conditions, including the optimizer, batch size, data augmentation strategies, optimized objective, and the number of samples seen during training.
For AMBIENT6K \cite{vasluianu2024towards}, we adopt their publicly available benchmark. 

Reading \cref{tab:quanti-test}, it becomes obvious that RLN$^2$ is a high-performance solution, improving over similar state-of-the-art solutions at all complexity levels, on both CL3AN and AMBIENT6K \cite{vasluianu2024towards}. 
While RLN$^2$-S largely outperforms RetinexFormer \cite{cai2023retinexformer}, at a lower computational cost, RLN$^2$-Sf improves over the perceptual performance of IFBlend \cite{vasluianu2024towards}, with a fraction of its required floating point operations.   

Against IFBlend \cite{vasluianu2024towards}, a strong recently proposed solution in the ALN field, RLN$^2$-L shows an advantage, especially in joint RGB-Frequency domains processing, with an $\approx$15\%  complexity reduction, while outperforming it. 

Another important observation is regarding the architectural core of the compared methods. Under both white-aligned and RGB direct lighting, ALN tends to favor solutions showing an extensive receptive field \cite{zamir2021multi, jing2021hinet}, rather than transformers optimized for local dependencies \cite{liang2021swinir, wang2022uformer, zamir2022restormer}, even when balancing to wider context blocks \cite{li2023grl}.
However, MPRNet \cite{zamir2021multi} and HINet \cite{jing2021hinet} are larger models, with an additional cost of approx. 63\%-87\%, compared to the highest complexity RLN$^2$-Lf variant. 
While state-state models prove computationally ineffective for the level of performance achieved, frequency-domain information access shows a consistent advantage, especially in terms of observed perceptual quality.   

\noindent\textbf{Qualitative Results:} In \cref{fig:qualitative-cle3n}, we provide for qualitative evaluation a set of samples representing a cluttered scene under the effect of non-homogeneous RGB direct lighting. 
Note the complex geometry and the wide RGB spectrum represented at the ambient lit image level. RLN$^2$-Lf shows improved behavior in terms of color normalization (row 1), superior restorations in highlights-rich areas, and better high-frequency shadow areas compensation (row 2). 
 
 \cref{fig:qualitative-ambient} provides for visual comparison equivalent renderings against the top performing state-of-the-art methods the AMBIENT6K \cite{vasluianu2024towards} benchmark. 
Consistently, the restored ambient normalized images are characterized by more natural colors, an increased amount of restored details, and geometrically correct renderings even in areas severely affected by underexposure (second row). Overcoming the challenges specific to non-homogeneous direct color lighting, RLN$^2$ shows consistent performance, confirming the advantages quantified through  evaluations provided in \cref{tab:quanti-test}. 

\begin{figure}
    \centering
    \includegraphics[width=\linewidth]{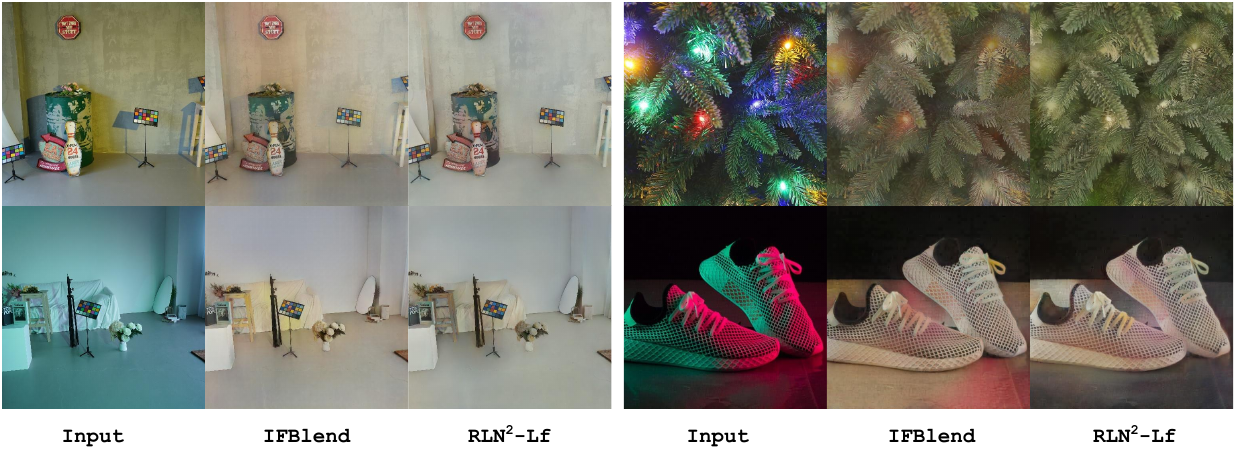}
    \caption{IFBlend\cite{vasluianu2024towards} and RLN$^2$-Lf on real-world images from LSMI \cite{kim2021large} (\emph{left}), or real product photography samples (\emph{right}).}
    \label{fig:fig-rw}
    \vspace{-5mm}
\end{figure}

\begin{figure}[t!]
    \centering
    \includegraphics[width=\linewidth]{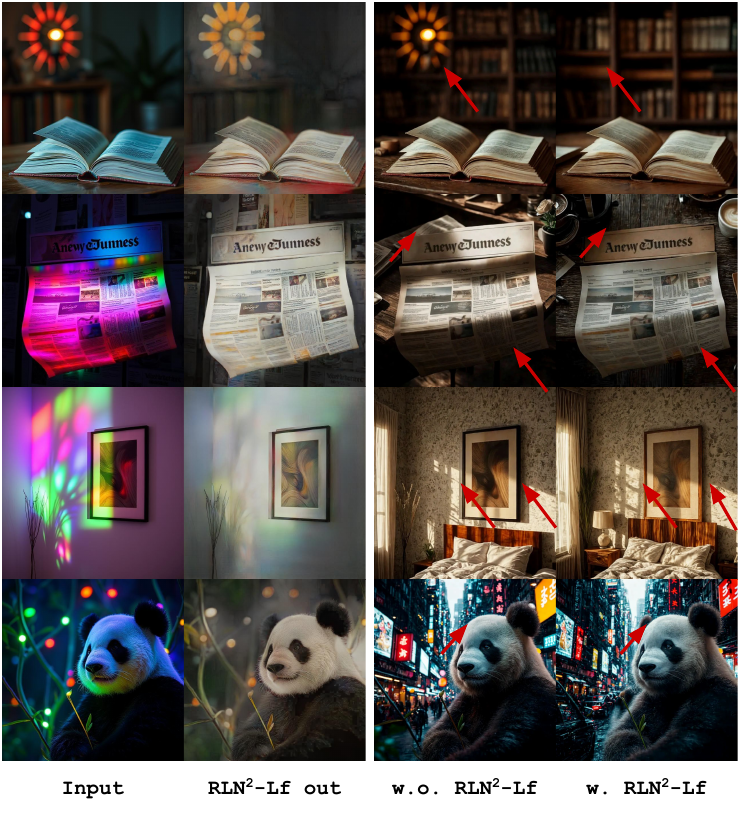}
    \vspace{-7mm}
    \caption{ALN as a preprocessing step before image relighting with foreground condition. Results rendered with ICLight \cite{fu2023guiding} for the input image (col. 3) vs. the RLN$^2$ output (col. 4).}
    \label{fig:fig-editing}
    \vspace{-7mm}
\end{figure}

To test RLN$^2$ on out-of-domain data, in \cref{fig:fig-rw} we compare IFBlend \cite{vasluianu2024towards} and our proposed RLN$^2$-Lf on inputs from LSMI \cite{kim2021large}, or complex lighting scenarios from product photography.
Our RLN$^2$-Lf shows increased robustness, with plausible rendered colors under uniform white ambient lighting conditions.  
Moreover, in \cref{fig:fig-editing} we deploy it on AI-generated images\footnote{\url{deepai.org/machine-learning-model/text2img}} representing various contents in challenging lighting environments. 
Applying our ALN solution as a pre-processing step in Neural Image Editing proves beneficial for IC-Light \cite{fu2023guiding}, with improved layer separation leading to less semantic discontinuities (rows 1, 2, and 4), and improves the rendered results quality in term of shadow casting (row 3) and foreground clarity (row 2).

\section{Conclusion}
\label{sec: conclusion}
In this work, we propose extending the current ALN study to the challenges added by multicolor and multi-intensity RGB direct lighting. We propose \textbf{CL3AN}, a high-resolution image database showcasing a multitude of materials under complex RGB lighting scenarios while providing white-light ambient-lit reference images using a professional photography setup.  We introduce RLN$^2$, a high-performance ALN algorithm setting a new standard for the ALN task, showing how alternative image representations can provide valuable control signals in feature refinement, or offer natural bypassing options for detail preservation. 

\noindent\textbf{Acknowledgments.} We thank the reviewers and ACs for their sustained effort and helpful feedback. This research was partially supported by the Alexander von Humboldt Foundation.

{
    \small
    \bibliographystyle{ieeenat_fullname}
    \bibliography{main}
}

\end{document}